# A Digital Shadow for Modeling, Studying and Preventing Urban Crime

Juan Palma-Borda, Eduardo Guzmán, and María-Victoria Belmonte

*Abstract*—**Crime is one of the greatest threats to urban security. Around 80% of the world's population lives in countries with high levels of criminality. Most of the crimes committed in the cities take place in their urban environments. This paper presents the development and validation of a digital shadow platform for modeling and simulating urban crime. This digital shadow has been constructed using data-driven agent-based modeling and Simulation techniques, which are suitable for capturing dynamic interactions among individuals and with their environment. Our approach transforms and integrates well-known criminological theories and the expert knowledge of law enforcement agencies (LEA), policy makers, and other stakeholders under a theoretical model, which is in turn combined with real crime, spatial (cartographic) and socio-economic data into an urban model characterizing the daily behavior of citizens. The digital shadow has also been instantiated for the city of Málaga, for which we had over 300,000 complaints available. This instance has been calibrated with those complaints and other geographic and socio-economic information of the city. To the best of our knowledge, our digital shadow is the first for large urban areas that has been calibrated with a large dataset of real crime reports and with an accurate representation of the urban environment. The performance indicators of the model after being calibrated, in terms of the metrics widely used in predictive policing, suggest that our simulated crime generation matches the general pattern of crime in the city according to historical data. Our digital shadow platform could be an interesting tool for modeling and predicting criminal behavior in an urban environment on a daily basis and, thus, a useful tool for policy makers, criminologists, sociologists, LEAs, etc. to study and prevent urban crime.**

## I. INTRODUCTION

Crime is one of the main problems in most urban environments, having a direct impact on citizens' quality of life. For this reason, governmental authorities try to deploy strategies to prevent, minimize, and palliate crimes. Predictive policing can be defined as the *collection and analysis of data about previous crimes for identification and statistical prediction of individuals or geospatial areas with an increased probability of criminal activity, to help developing policing intervention and prevention strategies and tactics* [1]. The most widespread approaches for predictive policing are based on what is called macro-emulation, which uses aggregated crime rates, demographic data, and other related factors through (mainly) statistical regression models and aims to discover crime spatial patterns. In urban environments, crimes are not evenly distributed but tend to concentrate around what are called hotspots [2]. Hotspots tend to persist over time, so their identification can be a fundamental element in the application of police methodologies and prevention strategies [3]. Consequently, one of the current lines for crime prevention is the identification of such hotspots with the help of techniques such as data analysis, data science, or even artificial intelligence (AI). As outlined in the routine activity theory (RAT) [4], criminal opportunities arise from routine activities of offenders, victims, and guardians. Crimes can only take place when an offender, a victim, and the absence of a "capable" guardian come together. For this reason, it is essential to be able to construct realistic urban crime models that can accurately represent the behavior of people in an urban environment in order to understand why and when crime occurs and to develop strategies to prevent it.

In recent years, the rapid advancement of emerging information sources like IoT, the enhancement of computing capabilities through cloud computing, and the evolution of analytical techniques such as AI and data science have progressively transitioned us from conventional static simulation models to the digital twin concept [5], enriching our comprehension of the environment. The term digital twin (DT) describes a virtual representation, i.e., the virtual entity (VE), of a physical system, that is, the physical entity (PE). Initially, this term was merely descriptive, but as soon as computational and information technologies evolved, it was possible to establish a bidirectional coupling between the PE and the VE. At present, there is no uniformity in the definition of a DT or in the understanding of the components that comprise a DT [6]. Despite this disparity, there is some agreement on what should be the elements that characterize a DT, i.e., a model, the data, and the synchronization between the PE and the VE. The

This research is partially supported by the Spanish Ministry of Science and Innovation and by the European Regional Development Fund (FEDER), Junta de Andalucía, and Universidad de Málaga through the research projects with reference PID2021-122381OB-I00 and UMA20-FEDERJA-065. (*Corresponding author: E. Gumán*).

J. Palma is with the Dpto. Lenguajes y Ciencias Computación, Universidad de Málaga, Málaga SPAIN (e-mail: jpalma@uma.es). E Guzmán is with the Dpto. Lenguajes y Ciencias Computación, Universidad de Málaga, Málaga SPAIN (e-mail: eguzman@ uma.es), M.V. Belmonte is with the Dpto. Lenguajes y Ciencias Computación, Universidad de Málaga, Málaga SPAIN (e-mail: mvbelmonte@ uma.es).

essence of DTs lies mainly in the exchange of data between the PE and the VE through bidirectional data connections. Additionally, a DT system should also include a set of services that enable the exploitation of data exchanged by the two twins in various ways. Examples of such services include, among others, dashboards for visualizing and displaying data in different formats and machine learning (ML) components to provide decision support and alerts to users. More specifically, the five-dimensional model of a DT [7] indicates that its main elements are the PE, the VE, the data, the connections, and the services that allow the exploitation of the data exchanged between the two twins. In addition, the term digital shadow (DS) is often used as a synonym for DT [8]. Kritzinger et al. [9], however, differentiate between the two terms based on the level of data integration. In DSs the data exchange between the digital and physical twin takes place through unidirectional data connections, which only map data from physical entities to digital objects, i.e., a change of state of the physical object leads to a change in the digital object, but not vice versa [9]. The DT would be at the highest level of connectivity, connecting both spaces automatically. Since 2003, when the term DT was first mentioned in the context of product lifecycle management [10], this concept quickly jumped to smart manufacturing and is now closely linked to the Industry 4.0 domain. However, specific DTs and DSs are now emerging in various areas such as smart cities, agriculture, or healthcare [11] [12]. A particular case of these digital artifacts is the urban digital twins (UDTs) or digital twin cities [13]. A UDT is a virtual representation of a city's physical assets, which uses data, data analytics, and ML to aid simulation models that can be updated and modified in real time as their physical counterparts change. In recent studies [14] [15], we can find works using DT and DS in the field of smart cities, such as urban planning, traffic, mobility and fleet management, healthcare, pollution control, sustainability, or energy. However, we have not found any DT or DS for police prediction or crime modeling.

Having appropriate technologies to model DTs or DSs is a challenge. In this regard, agent-based modeling and simulation (ABMS) is a useful approach to represent and study complex systems. Since the 90s, it has been one of the most prominent fields of social sciences research, and it is used as a technique to simulate and understand a great diversity of social phenomena. In general, an agent-based model is *composed of individual entities that have autonomous behaviors and are different from one another, having diverse characteristics and behaviors over a population* [16]. Such a model typically consists of a set of autonomous agents, characterized by a set of attributes and a collection of specified behaviors, connected to other agents and their environments by certain relationships [17]. These agents, their environment, and their relationships, therefore, conform a system with clearly defined boundaries, inputs, and outputs. ABMS, in contrast to other simulation methods, is a bottom-up approach [18].Therefore, it helps to understand both agents' and aggregate behavior and explains how these behaviors lead to large-scale outcomes and emergent phenomena [19]. From this perspective, researchers can model how macro-phenomena emerge from behavior at the micro-level among a heterogeneous set of interactive agents [20].

Communication technologies have become ubiquitous, allowing the collection of data from a wide variety of sources. When made available for ABMS, a large volume of data can beneficially reveal hidden insights and behavioral patterns of humans. Consequently, some of the current challenges and problems in ABMSs are often related to an increase in used and produced data. Through the combination of ABMS with techniques coming from fields such as Data Science or ML, it may be possible to explore spaces and design options with a new focus called data-driven ABMS (DABMS) [21]. This approach makes agents' behavior more realistic incorporating more assumptions on agents' routines, based on empirical data, and calibrating and validating them against real-world data [16]. DABMS also allows modeling environments fed (continuously) with real-time data, which facilitates the study of optimal resource deployment strategies for real environments. This use of DABMS fits with the current concept of DT or DS [6] [7] [22] and enables professionals from other fields to optimize their decisions on real-world problems. So, the use of DABMS would allow exploring different crime theories [23] and their consequences on the urban environment, where it is generally unfeasible to conduct real-world experiments. Until relatively recently, software simulators that emulated real environments used limited datasets, obtained from the real world or generated synthetically, and provided their results off-line (or asynchronously), which had to be interpreted by stakeholders who in turn translated those interpretations into policies or measures to be applied in the real world. However, with recent advances provided by DTs or DSs, now the simulator and model can essentially be a digital shadow that continuously evolves as it is fed with new real-world data [5].

In this paper, we present a DABMS-based DS platform to predict the daily criminal behavior in an urban environment that also works as a workbench environment for policy makers, criminologists, sociologists, etc. In our DS, the agent decision-making strategies rest on a large amount of real data, which allows a realistic model of the agents' behavior, combined with several well-founded criminological theories, which will be specified in the paper. Regarding the importance of using a realistic spatial environment, our approach models crime at grid cell level, which is an aggregation of several streets. Gridded cells are the most popular way of representing simulated environments [22] [24] [25] [26]. The DS has been instantiated, calibrated, and validated for the city of Málaga (Spain), using the full crime database of the city provided by the national police force and collected between 2010 and 2018. The results suggest that our modeling proposal is suitable for the characteristics of this city. According to the performance results and to the metrics commonly used in the field of predictive policing, the generation of simulated crimes through our model of the city of Málaga is aligned with its real crime data.

Next, the contributions of this paper are summarized below:

- We first propose a systematic approach and a soundness methodology, based on DABMS, to build urban crime simulation environments. Our proposal uses crime data combined with spatiotemporal and socio-economic characteristics of the citizen population. Accordingly, agents' decision-making strategies are based on real data and sound criminological theories. This allows for a more realistic modeling of the agents' behavior and their likelihood of committing a crime.

- The low granularity of our approach for spatial representation (we use a grid cell size lesser than 200 $m^2$) is suitable for analyzing crime in urban environments, having enough precision to identify hotspots within neighborhoods or urban districts.
- Our DS platform has been tested for modeling the daily criminal behavior in the city of Málaga (Spain), for which we had the full crime database of the city between 2010 and 2018 and socio-economic and geographic information of the city at our disposal. The calibration and validation have been supported by the full database of real crime data, with over 300,000 reports.
- To validate the performance of the model of Málaga, a large number of well-specified metrics were used to facilitate easy comparison of this work with others in the field and to differentiate it from a significant proportion of existing state-of-the-art papers. Our proposal achieves good predictive values, according to the metrics used in predictive policing. The results reveal that the model can provide realistic crime prediction.
- The model can easily be scaled up to serve as a test bed for field practitioners (LEAs, criminologists…) to perform different tasks, including crime rate prediction, hotspot detection, patrol strategy optimization, crime network analysis, etc.

The paper is structured as follows: The next section summarizes the background on simulation modeling techniques for predictive policing, specially focusing on ABMS. Section III provides a snapshot of the DS platform architecture and shows the process that must be followed in the instantiation, calibration, and validation of the DS for a particular urban environment. Section IV is devoted to data collection and in particular to explaining the data on the city of Málaga we have used in this work, that is, the reports on crime, the cartographic information of the urban environment, and the socio-economic data of its population. Section V mainly describes how the design of our DS has been performed, highlighting the construction of the urban environment, their agents, the global parameters, and the algorithms behind the simulation. The theoretical fundamentals and behavior of the model during the simulation are approached in Section VI. For instance, this section defines concepts such as the criminal power of an urban area or the likelihood of a citizen committing a crime, which are relevant to describing the dynamics of crime in the city. Section VII explains how some not directly observable model parameters were calibrated and how the calibration process was validated. Section VIII analyzes and discusses the model's contributions and outlines some limitations, and, finally, section 9 draws some final conclusions and future works.

## II. BACKGROUND

Simulation modeling provides a strategy for capturing dynamic interactions among individuals taking place at the micro-level and their relationship to macro-level patterns. It involves the creation of a simplified representation of social phenomena [27], and it is suitable for representing non-linear relationships present in dynamic and complex interactions. In addition, simulating crime patterns contributes to the understanding of urban crime in a spatial environment [28]. In this sense, agent-based models are especially useful as computational approaches for modeling human beings to study social phenomena [29]. ABMS techniques have been applied to the study of crime in a wide variety of setups, including theory testing [30], testing of prevention strategies [24], and predicting crime patterns [31]. Brantingham & Brantingham [32] inspired much of this interest with a seminal article in which they described how ABMS could explain opportunity. Later, in their review, Groff et al. [23] identified 45 ABMSs of urban crime. One advantage of simulation techniques is that they make it easier to perform experiments that would be unfeasible in real life since field experiments are expensive and hard to develop. An ABMS-based experiment allows us to examine the same units under different experimental conditions. Also, it has the added rigor of exploring potential outcomes across a series of trials. For instance, ABMSs can be used by LEAs to test the effect of patrolling strategies on crime [24] [25] or by urban planners to assess how changes in urban design may affect the development of crime.

The first ABMS-based approaches studied the dynamics of crime without including spatiotemporal characteristics [22] [33], and only 33% of crime simulation proposals were spatially explicit [11]. For example, Winoto et al. [34] developed an agent-based model that applies a choice-theoretic approach in which agents decide whether or not to commit a crime only according to their perception of the costs and benefits. Nowadays, there is a growing trend that sets up the relevance of using realistic spatial environments for modeling [35] [36]. For large areas, grid cells are the most popular choice for modeling urban environments [22] [24] [25] [37]. Grid cells divide the general space into equal size units, so that each cell may contain different types of information, e.g., number of people and crimes, geographical information (residential, transportation-related, commercial), etc. However, the size of these units is sometimes arbitrary, which can call into question the validity of the results. Additionally, placing crime where it actually occurs may be an important issue in crime prediction, but [38] reveals that the results for prediction at street segment level are comparable to models in literature based on larger units of analysis such as grid cells. Other approaches for modeling spatial environments are the transportation networks in the forms of nodes and edges. These are also a popular election due to their efficient implementation and accurate routing [28] [39]. Nevertheless, the interactions and movements of agents are constrained by the nodes and edges and also introduce too many crime opportunities when people congregate at the same points rather than being distributed in a random space [40].

TABLE I

SELECTED PROPOSALS ON URBAN CRIME MODELING AND SIMULATION FROM THE LAST DECADE

| Authors | Crime data volume | Study area | Spatial environment | Technique | Crime theories | Perfomance |
|---|---|---|---|---|---|---|
| Wang et al. 2012 | 1,795 incidents | USA, Virginia, Charlottesville | Grid cells of 32m × 32m | Spatio-temporal generalized additive models | Not specified | Hit rate 3% = 50% |
| Devia et al. 2013 | 4 years of crime reports, Santiago. Unknown for Vancouver | Chile, Santiago de Chile, and Canada, Vancouver | Grid cells of unknown size | ABMS | Routine activity theory | None |
| Adepeju et al. 2015 | 9 month of crime reports | USA, Illinois, South Chicago, and England, London, Camden | Grid cells of 250m × 250m | Prospective hotspot<br>Self-exciting point process<br>Prospective kernel density estimate<br>Prospective space-time scan statistic<br>Predictive selective random algorithm | Not specified | Accuracy from 39.8% to 91.5%<br>PAI 20% = 4.58 |
| Rummens et al. 2017 | 163,800 crimes | Belgium, city > 250,000 people | Grid cells of 200m × 200m | Ensemble model<br>Synthesizing logistic regression<br>Neural network model | Not specified | Precision<br>Home burglary day = 57.87%<br>Home burglary night = 52.02% |
| Weisburd et al. 2017 | — | USA, mix of the 20 biggest cities | Grid cells of 14.32m × 14.32m | ABMS | Routine activity theory<br>Rational choice theory | None |
| Kadar et al. 2019 | 3 years of crime reports | Switzerland, Aargau canton | Grid cells of 200m × 200m | Machine learning models | Crime pattern theory<br>Social disorganization theory<br>Near repeat victimization theory | AUC = 0.844<br>Hit rate at 5% coverage = 51%<br>Hit rate at 20% coverage = 74.4% |
| Lee et al. 2020 | 1,018,864 calls of services, Portland; 164,795 crimes, Ohio | USA, Oregon, Portland, and USA, Ohio, Cincinnati | Grid cells of 152m × 152m | Forecasting model using Poisson probability of crime for each month | Population heterogeneity | PEI = 82.3<br>Accuracy = 56.2% |
| Gong et al. 2021 | 4,479 crimes, 495 of them violent crimes | USA, Virginia, Hampton | Small GIS regions connected by real roads | ABMS | Rational choice theory<br>Routine activity theory | No standardized metric:<br>Accuracy between 257m to 901m to real hotspots = 82% |
| Zhu et al. 2021 | 76 robberies | USA, Louisiana, East Baton Rouge Parish | Grid cells of 500m × 500m and road points inside the cell | ABMS | Routine activity theory | Hit rate = 63%<br>PAI 7.2% = 8.7295<br>FAI = 1.09 |
| Rosés et al. 2021 | Training / Test<br>16,413 / 1,303 robberies | New York City, New York, USA | Street segments | ABMS | Crime pattern theory<br>Routine activity theory<br>Broken windows theory | PAI 3% = 10.25<br>PAI 20% = 4.51 |
| Escobar et al. 2023 | 398,103 crimes | Mexico, Jalisco, Guadalajara | Grid cells of unknown size and a streets layer | ABMS | Crime pattern theory<br>Routine activity theory | None |
| Hakyemez et al. 2023 | 24,151 robbery and theft incidents | USA, Illinois, Chicago | Graph | Graph Wavenet, STGCN, and LSTM | Not specified | Total crimes PAI 20% = 3.98.<br>Daily theft PAI 3% ≈ 13.3.<br>Daily robbery PAI 3% ≈ 6 |

Most ABMS proposals implement theory-driven simulations. According to [26], the RAT, the rational choice theory (RCT) and the crime pattern theory (CPT) are the main crime theories used to define crime incidents and offender behavior. Among them, RAT is the foundation of most crime simulation proposals [24] [25] [26] [27] [38] [41] [42]. RCT suggests that offenders exhibit a rational behavior by considering specific aspects of a situation before deciding whether to commit a crime [40] [25]. According to CPT, crimes occur when the activity space of victims and offenders collide [43]. For example, Malleson et al. [44] used theoretically informed agent decision-making processes, such as the behavioral framework called PECS [29]. Birks et al. [45] studied the impact that several theoretical propositions, i.e., RAT, RCT and CPT, have had on offending patterns by means of a series of simulated experiments. However, a shortcoming of these approaches is that decision-making by agents is largely based on theories and concepts, and only in some cases is supported by limited empirical data. Furthermore, some proposals model a city without the presence of any kind of guardian, which could be unrealistic. In this sense and according to RAT, the presence of "capable guardians" may intervene or deter offenders by their very presence before any crime has been committed.

In Table I, we have summarized several proposals on urban crime modeling and simulation from the last decade. The table is not intended to be exhaustive but to illustrate and compare the main characteristics of these studies, such as volume of crime data, study area, spatial environment, techniques used, criminological theories included to define agents' behaviors, and spatial conditions of crime and performance. Regarding the technique employed in modeling the environment and the interactions, most of the works included in the table correspond to ABMS in order to facilitate a comparison with our approach. Moreover, some others reviewed proposals [46] [47] use ML models. Zhang et al. [48] also compare six ML algorithms for predicting crime hotspots. ML-based proposal, especially neural networks or ensemble methods [46] [49] [50], are high-performance prediction models due to their ability to process non-linear relational patterns in data. Additionally, they can handle very high-dimensional data and extract the characteristics of the data, but their interpretability and traceability are limited. In predictive policing, explaining predictions is an essential concern of decision-making as it allows users to trust and use predictions in a right and effective way. For this reason, the building of intrinsically interpretable decision models has been a main objective of our work, aligned with the Explainable Artificial Intelligence (XAI), which proposes making a change toward a more transparent AI [51].

Focusing on ABMS proposals included in the table, several studies have highlighted that ABMS is a potentially powerful tool for hotspots policing. Weisburd et al. [25] draw on agent-based models to examine the crime prevention impacts of hotspots policing on street robbery patterns in a small urban area. However, the approach presents some shortcomings: the simulation is built not on an entire and real city but rather on a fictitious small urban area, with a population density and citizen-police ratio that are consistent with the averages for the top 20 populated cities in the United States. The simulation model is not validated with real data, and validation is of particular importance since the artificially generated data must not necessarily be representative of the real phenomena they are intended to reflect. Devia et al. [24] present a model that generates artificial street-crime data that can be used to test different policing strategies in a virtual environment. As indicated in this work, the validation results were quite questionable, so the generated data were not a faithful reflection of the real crime hotspots. This work also did not focus on large urban areas, but on medium-sized city centers (an area of 3 × 2 km). Gong et al. [40] establish an ABMS-based model that incorporates both criminological theories and geographical information, as well as defining different behavioral profiles for citizens. However, the validation presented is rather limited, utilizing only proximity of predicted crimes to actual hotspots as metrics without considering their distribution. Escobar et al. [42] propose another model aiming to predict crime patterns, ultimately inferring the existence of hotspots within a city where crime is concentrated, but their validation is lacking as they do not provide metrics to support their findings or enable comparison with other studies. Zhu & Wang [26] establish a scalable agent-based model for simulating urban crime, which enables the evaluation of various police strategies and includes clear metrics for comparison with other models. Still, due to its limited sample size of only 76 robberies, the study lacks sufficient data for robust validation, and this compromises the reliability of its conclusive results. Roses et al. [38] propose an ABMS with a robust validation stage, incorporating spatial or mobility data to construct the model. Nevertheless, the extensive amount of data, particularly the presence of LBSN data, renders its application impractical for the vast majority of cities lacking access to such dataset. Later, in the discussion section, we will approach the comparison, mainly regarding the performance, between our proposal and the works summarized in Table I. 

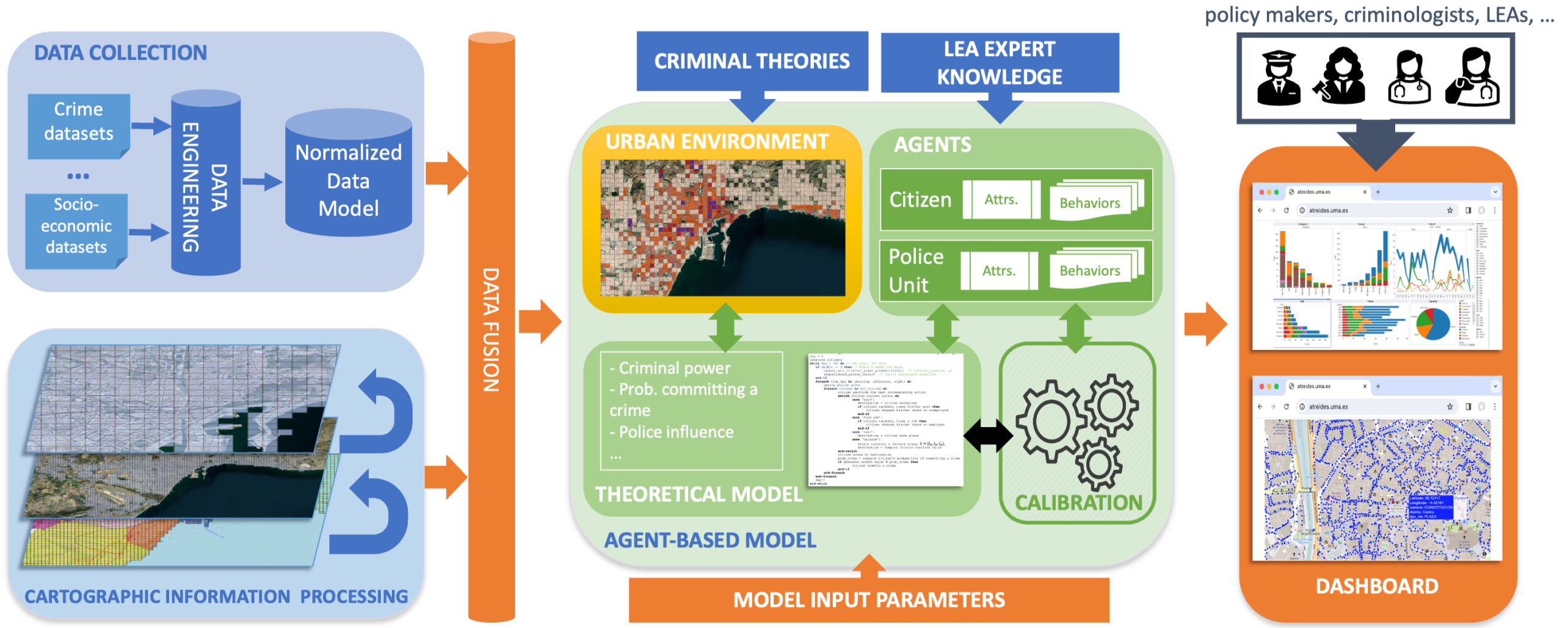

**Fig.1**. Overview of the architecture of our DABMS-based DS

## III. MODEL OVERVIEW

Fig. 1 illustrates the architecture of our DS, which consists of four different modules. The first two modules perform the input data management in order to prepare the information required by the DS to operate as a data-driven architecture. The first one, responsible for the data collection, combines and converts all the input data sources into a unified and normalized data model. The main data source is the information on crimes, containing the reports received and collected by LEAs or available through open data repositories. This information is also combined with the available socio-economic data on the urban environment selected to be modeled, such as the population per district or area, the rate of unemployment, etc. All these data are processed through a stage of knowledge engineering where they are fixed, categorized, imputed, etc. to be normalized and inserted into a data model. Parallelly, cartographic information of the urban environment is also processed to transform the environment into a bidimensional space of cells, each one mapping a small urban sector. If available, the information of districts or areas in which the urban environment is structured has been included in the grid. Posteriorly, cartographic information and normalized data model are merged into a single representation where the data model is enriched with the geographical information, and also the attributes of the cells are populated with socio-economic and geographic data of the urban sector they represent. The central part of the architecture is the agent-based model, which is fed with the data generated after the fusion process. Those data are used to parameterize the simulation agents and the internal parameters of the model. The behavior of the simulation agents and the daily routines of the citizens and offenders have been formalized accomplished in terms of well-founded criminal theories and the expert knowledge of the LEA members and criminologists who are involved in the research project behind this DS. The DS needs to be calibrated, and to this end, the normalized information is used according to a set of global input parameters. Finally, once calibrated, the dashboard of the DS allows LEAs, criminologists, policy makers, etc. to study the trends on the urban crime globally or in terms of different offenses, districts, etc., analyze hotspots, test different police deployment strategies, etc.

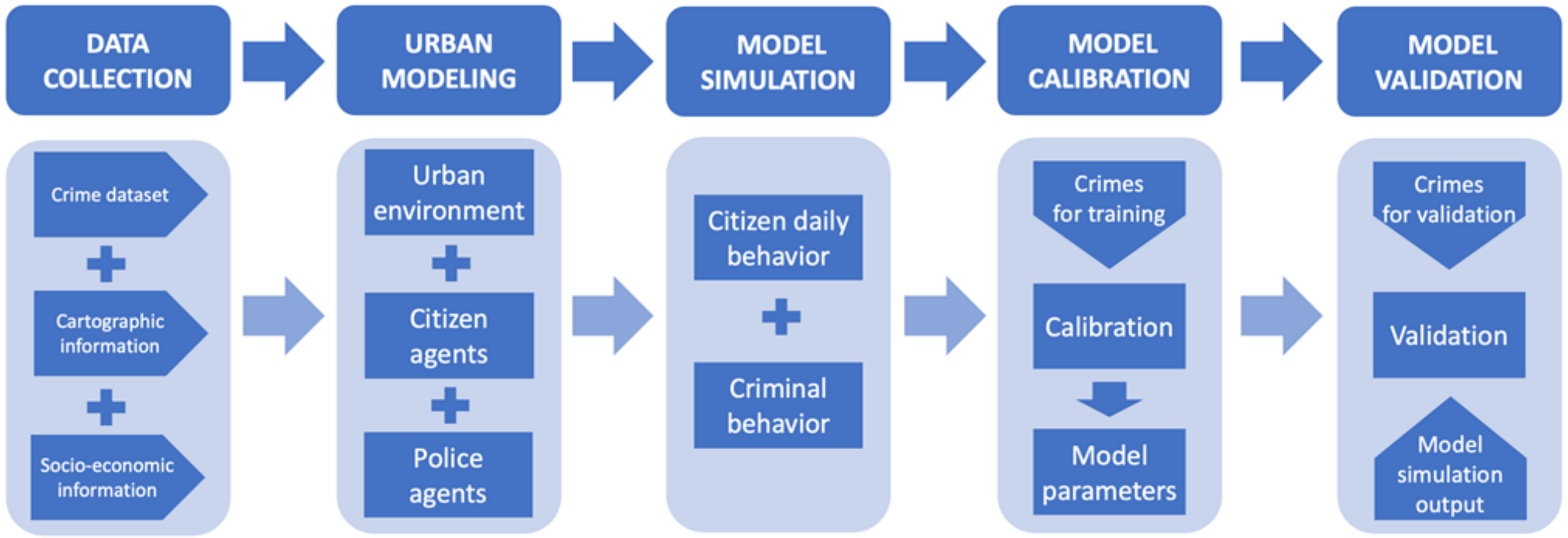

**Fig. 1.** Methodological flowchart followed in the construction and validation of our DABMS-based DS

Fig. 2 shows a flowchart that summarizes the methodology followed in the instantiation, calibration, and validation of the DS for a particular urban environment. As mentioned above, the first two modules of the architecture oversee the processing of the input information; all tasks performed in those modules correspond to the data collection step. Next, in the urban modeling step, the agent-based model is instated and populated. The model simulation step provides the dynamic behavior of the agents and the global urban model. Finally, the model is parameterized with global parameters, and those internal latent parameters, not directly observable, have to be calibrated splitting the crime dataset in a training and validation subset to accomplish first the calibration and later the model validation. Once validated, policy makers, LEAs, criminologists, sociologists, etc. can use the dashboard of the DS to explore, analyze, and test hypotheses on urban criminal behavior.

In the next sections, we will describe the steps represented in Fig. 2 and illustrate them with the case study of the city of Málaga. This city has been selected for two main reasons: we had the full crime database for 2010–2018, and the design and construction of the DS have been performed in the framework of a research project where technicians, criminologists, policy makers, and experts from the LEAs that operate on that city are involved.

## IV. Data Collection

### A. Crime Data

The main input data used by the DS platform is the information related to the crimes that take place in the urban environment. The platform requires at least data related to the typology and the spatial and time location of the crimes, that is also the information that can be found in most open data crime repositories. In the case of Málaga, in principle, we had access to reports ranging from lost property reports, such as loss of legal documentation (e.g., ID card, passport, etc.), to minor offenses, such as wallet thefts, or serious crimes such as sexual abuses. These reports are filed by citizens, and concerning this work, each qualified report is thoroughly considered. This significantly reduces the potential risk of bias that may exist on the part of the police. The original datasets were taken directly from the police database in CSV format and separated into different files per year. The same report was often found in one or more rows of each dataset; thus, a first pre-processing stage of information identification and grouping was necessary to give them a proper semantic meaning. As a result, we obtained a set of reports, structured into the following 5 major information blocks:

- *Event*: event triggering the report and the place and date where it was committed.
- *Responsible Parties*: people who carried out the event.
- *Implicants*: victims or witnesses of the event.
- *Objects*: elements that were used to commit the crime or that were part of stolen goods.
- *Acting Patrol*: police unit in charge of collecting the report.

The initial total number of reports was around 800,000, but only 549,336 belonged to the city of Málaga. Fig. 3 depicts the classification of those crimes or reports provided by the police and circumscribed to the city. Note also that information regarding responsible parties, implicants, and/or objects was often missing (> 90% of null values), and data on the police units who reported the crime were removed since they did not contain useful information for our studies. For this reason, only the *event* block and the place and date where it was committed were considered. Finally, due to the nature of our work, all those reports that either did not represent any crime (i.e., *no offense*, 157,998 reports) or were outside the scope of the city (reports registered in the province of Málaga but occurred in other municipalities, that is, 14,601 crimes) were removed. At the end of the whole process, we kept data on 376,737 offenses in the city of Málaga, corresponding to the categories of *Infraction*, *Offense*, *Misdemeanor*, and *Minor Offense*. Within the selected categories, the variety of offenses and the classification made by police officers make it difficult to draw a general parallel between them and their legal and police definitions in other countries. Even so, as can be seen in Fig. 4, over 40% of the crimes in Málaga are property crime (theft, burglary, vehicle theft, and robbery). Drug-related crimes represent up to 23% and the most serious offenses, such as sexual abuse, murder, or kidnapping combined, represent not even 0.5% of the city's crimes.

## *B. Cartographic information*

Our DS platform requires information on the street map corresponding to the urban environment that wants to be modeled. In the case of Málaga, we first retrieved the full list of the roads and then the cartographic information of the street map provided by the open data of the city (https://datosabiertos.malaga.eu/). All this information was merged, but previously, a processing stage was required due to the errors found in the city street map. The final structure of the cartographic information used in this work is shown in Table II, which consists of four fields: 1) the road typology abbreviation using the shortening of the actual terms in Spanish, i.e., square (PL, PZLA), street (CL), avenue (AV), etc.; 2) the full name of the road; 3) the district field that refers to the council district number; 4) the street buildings (the geolocated point list is in EPSG:32630[1] format).

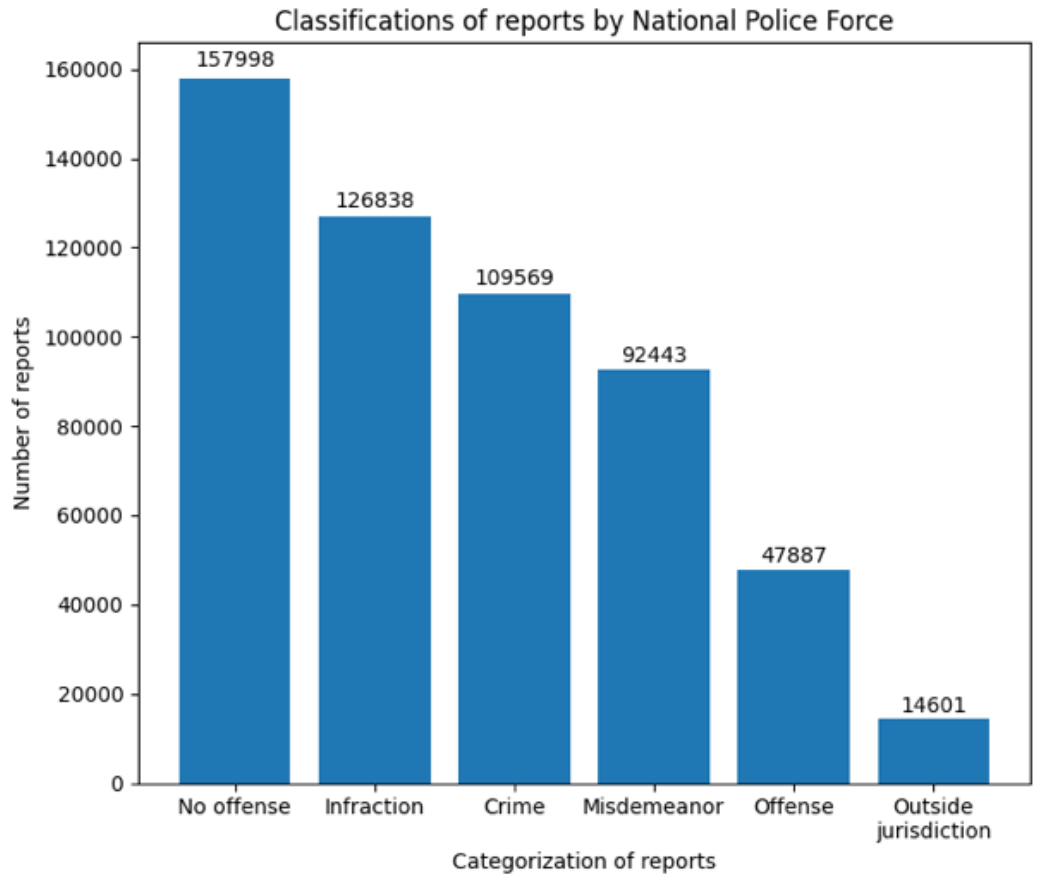

**Fig. 2.** Classifications of reports by the National Police Force in the city of Málaga

## *C. Socio-economic information*

In order to simulate the daily behavior of an urban environment and its inhabitants, the main socio-economic information used is its total current population and the percentage of population per district, since it helps to distribute citizen agents in a realistic way. Other relevant data useful for simulation are the percentage of unemployed or employed people. If some of this information is not available, it can be included as an input parameter of the model simulation. In the case of Málaga, we obtained the census information of the city by districts, shown in **¡Error! No se encuentra el origen de la referencia.**, and the average unemployment rate of the city, around 30.86%. Ideally, it would have been preferable to obtain the unemployment percentage by city district, but unfortunately such information was not available. This information has been obtained from the Spanish National Institute of Statistics. It should be mentioned that the choice of unemployment value is the average unemployment rate for years 2010–2017 in the province of Málaga and the actual value for the year 2014.

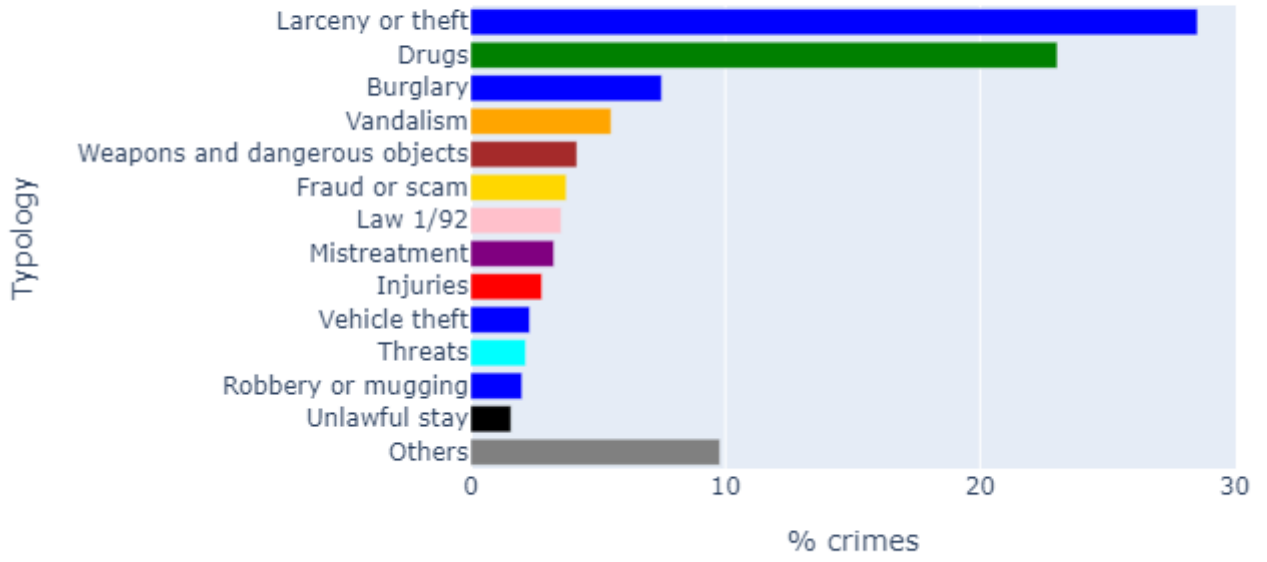

**Fig. 4.** Aggregation of crimes by the National Police Force in the city of Málaga.

## *D. Socio-economic information*

In order to simulate the daily behavior of an urban environment and its inhabitants, the main socio-economic information used is its total current population and the percentage of population per district, since it helps to distribute citizen agents in a realistic way. Other relevant data useful for simulation are the percentage of unemployed or employed people. If some of this information is not available, it can be included as an input parameter of the model simulation. In the case of Málaga, we obtained the census information of the city by districts, shown in Table III, and the average unemployment rate of the city, around 30.86%. Ideally, it would have been preferable to obtain the unemployment percentage by city district, but unfortunately such information was not available. This information has been obtained from the Spanish National Institute of Statistics. It should be mentioned that the choice of unemployment value is the average unemployment rate for years 2010–2017 in the province of Málaga and the actual value for the year 2014.

TABLE II
STRUCTURE OF THE CITY'S GEOGRAPHIC INFORMATION USED IN THIS WORK

| Field | Description | Type |
|---|---|---|
| Road abbreviation | Abbreviation of the road type | String |
| Road name | Name of the road | String |
| District | Number of the district | Int |
| Road buildings | List of geolocated road buildings using EPSG:32630 | List<Pair<Int, Int>> |

[1] EPSG:32630 is a specific coordinate reference system (CRS) within the EPSG Geodetic Parameter Dataset. It corresponds to the "WGS 84 / UTM zone 30N" projection, which is commonly used for mapping and navigation in the northern hemisphere, particularly in parts of Europe, Africa, and Asia; its units are meters.

Finally, through the data fusion process, the cartographic information of the streets is cross-referenced and related to the information of the crime dataset. Furthermore, each cell of the grid is labeled with the corresponding socio-economic information available and the district or area it belongs to. For the case of Málaga, crimes that could not be geolocated were eliminated, so that in the end we had a total of 304,125 geolocated crimes in the city. Fig. 5 shows a snapshot of the report distribution regarding the approximate location where the crimes took place.

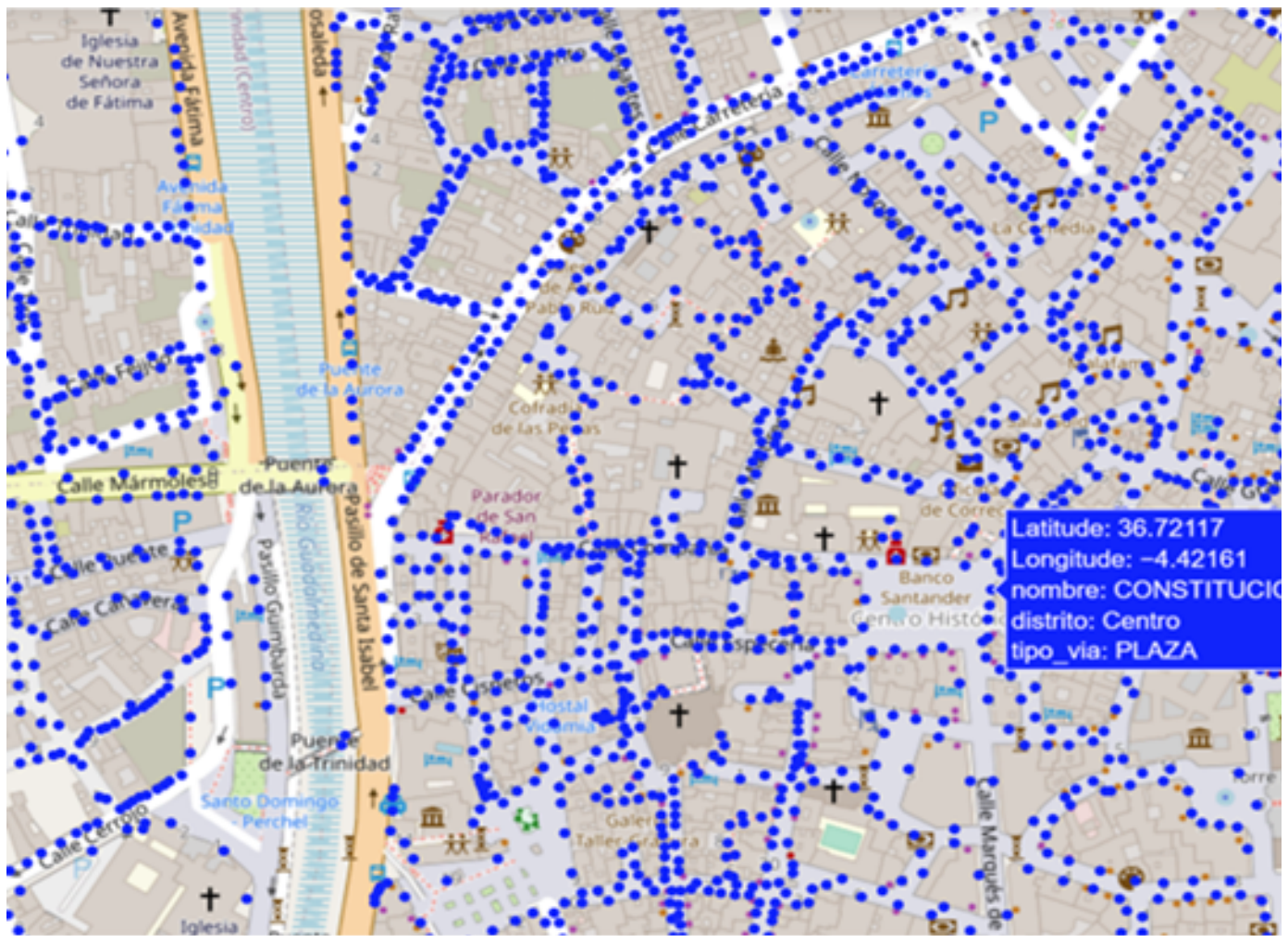

**Fig. 5.** A snapshot of the crime geolocation in the urban environment.

## V. THE URBAN MODEL

Our urban model can be expressed as a tuple (E, C, P, T) composed of a set of cells representing the urban geographical environment (E), a set of citizen agents (C), a set of police units (P), and a time dimension (T). The first subsection shows how the urban geographical environment, where simulations take place, has been modeled. In this sense, two dimensions have been considered for this purpose: the spatial dimension, by constructing a simplification of the urban environment where crimes are committed, and the time dimension, representing the daily dynamic behavior of the city. This second dimension is described in the second subsection. The roles of citizen agents and the police units that take part in the simulations are outlined in the third subsection. The fourth subsection enumerates the model input parameters and how we characterize the probability of an offense taking place in certain location of the urban environment.

### *A. The urban environment model*

The first dimension of our model consists in the construction of a spatial model of the city. Probably the most widespread and widely used mechanism to generate an urban geographical environment for simulation is a two-dimensional grid defined as a set, $E = \{e_{1,1}, e_{1,2}, \ldots, e_{n,m}\}$, where each $e_{x,y}$ represents a cell of the grid placed at coordinates ($x$, $y$). This type of scenario exhibits two main characteristics: 1) Even though most available crime datasets provide the approximate area where the crimes took place, this information does not tend to be very precise, as it only consists of the name of the road but not the exact point where the crime was committed. Therefore, building fine-grained environments would only introduce much noise in simulations. Dealing with larger areas greatly reduces the error in terms of the precision of the areas where crime occurs. 2) The hotspots identification for crime study and analysis purposes does not require a high degree of specificity, i.e., a corner or a part of a street. Identifying slightly larger areas, such a set of streets, a small part of a neighborhood, etc., is enough. However, it can be used for slightly larger areas, such as a set of streets, a small part of a city neighborhood, etc. For this purpose, to generate the city grid and, therefore, an environment for the simulation, the urban area to be gridded needs to be first chosen. In our DS platform, the dimensions of the square cells are recommended to be around 200 m$^2$, i.e., a space small enough that could be crossed on foot from one end of a cell to another in around 5 minutes, so that this area and another adjoining places could be covered by a police unit. In the case of Málaga, the area was gridded on a map of 40 to –40 from north to south and of 64 to –64 from east to west. Figs. 6 and 7 shows how this gridding process was performed. Each square has dimensions of approximately 195 m on a side, comprising around 38,000 m$^2$, i.e., approximately 3.8 hectares. Regarding the shape of the grid, note that Málaga is a seaside town, which is also located right on a corner of the coast. It leaves the city with the shape of the upper left corner of a square. As a consequence, if a square (or even a rectangle) were used to represent each cell of the city, it would include a larger number of inaccessible places (e.g., sea, untraveled natural places, and areas belonging to other municipalities). The areas of interest for the urban environment modeling were contained in the polygon surface, but, at the same time, the areas not belonging to the habitable area of the city were minimized.

TABLE III
STRUCTURE OF THE CITY'S GEOGRAPHIC INFORMATION USED IN THIS WORK

| **District name** | **District number** | **Current Population** | **Percent. of population** |
|---|---|---|---|
| Carretera de Cádiz | 7 | 115,391 | 20.16 |
| Cruz de Humilladero | 6 | 86,520 | 15.12 |
| Centro | 1 | 82,193 | 14.36 |
| Bailén – Miraflores | 4 | 60,604 | 10.59 |
| Este | 2 | 57,364 | 10.02 |
| Ciudad Jardín | 3 | 36,662 | 6.41 |
| Teatinos – Universidad | 11 | 35,452 | 6.20 |
| Palma – Palmilla | 5 | 30,727 | 5.37 |
| Puerto de la Torre | 10 | 29,581 | 5.17 |
| Churriana | 8 | 19,393 | 3.39 |
| Campanillas | 9 | 18,380 | 3.21 |
| **MÁLAGA** | | **527,267** | **100.00** |

Once the city grid is designed, it has to be combined with the street map of the urban environment, as well as a map outlining the borders of the city's districts. This integration needs to be subsequently reviewed to ensure that there are no errors in the representation of the districts (see Fig. 8). In addition, each cell has to be characterized by three attributes:

- *district*, that is, the name of the district or area where the cell is located;
- *habitability*, if there are buildings registered at that point in the city;

- *walkability*, i.e., the cell is neither sea, river, nor part of other nearby municipalities.

*B. The time dimension*

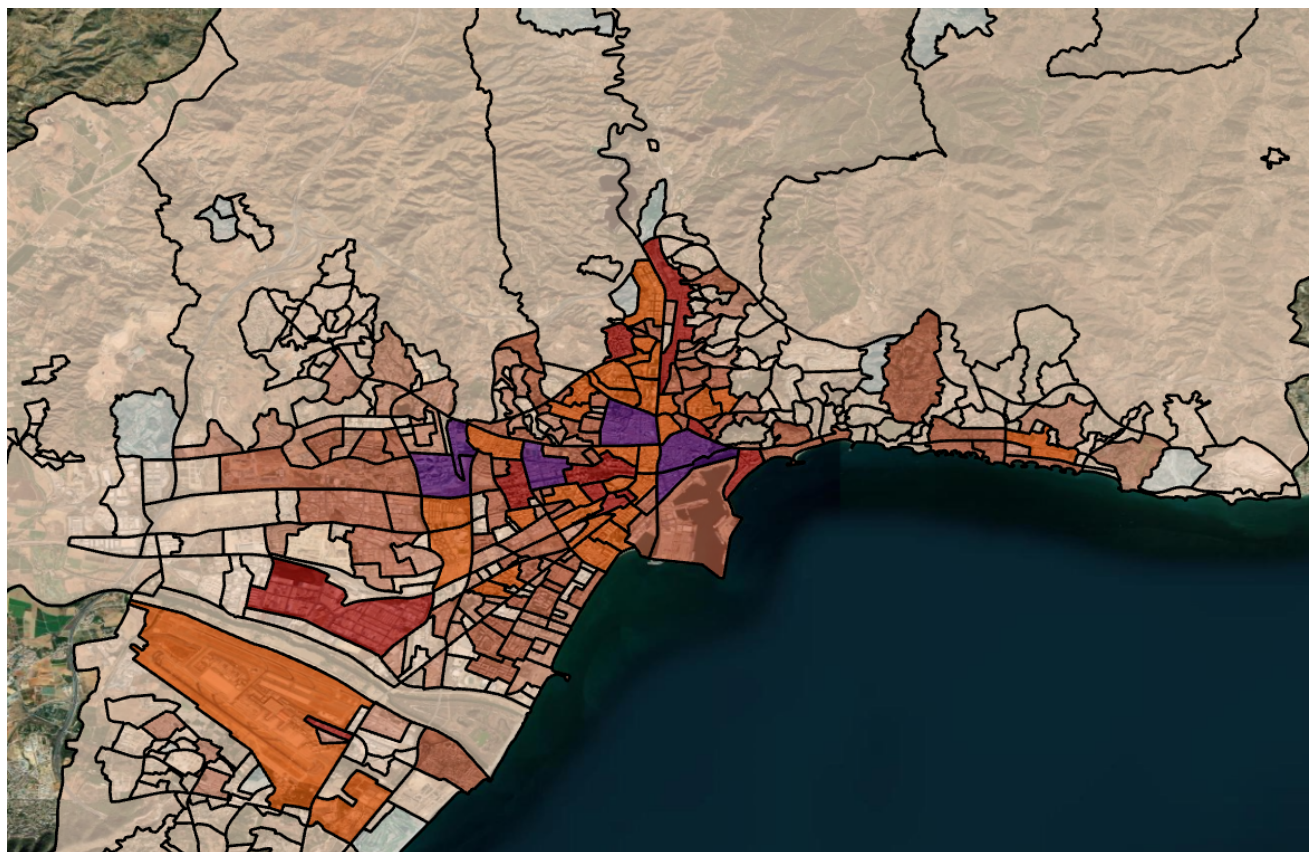

**Fig. 6.** Map of the city of Málaga

Once the environment is described at the spatial level, it is necessary to define how time will be represented in the simulations. The simulations are divided into days T = ($t_1$, $t_2$, …, $t_n$). To simplify the model, each day ($t_i$) is divided in turn into three consecutive time periods of 8 hours each: morning ($t_i^M$), afternoon ($t_i^A$), and night ($t_i^N$). This division corresponds with the time periods used by police officers to distribute their patrol shifts. It also allows us to simplify the model by avoiding the need to include a pathfinding algorithm to move the officers from one cell to another, since we assume that they have enough time to move from one point to another in the city in eight hours. Thus, each simulation changes their state from one time period to the next (e.g., morning to afternoon).

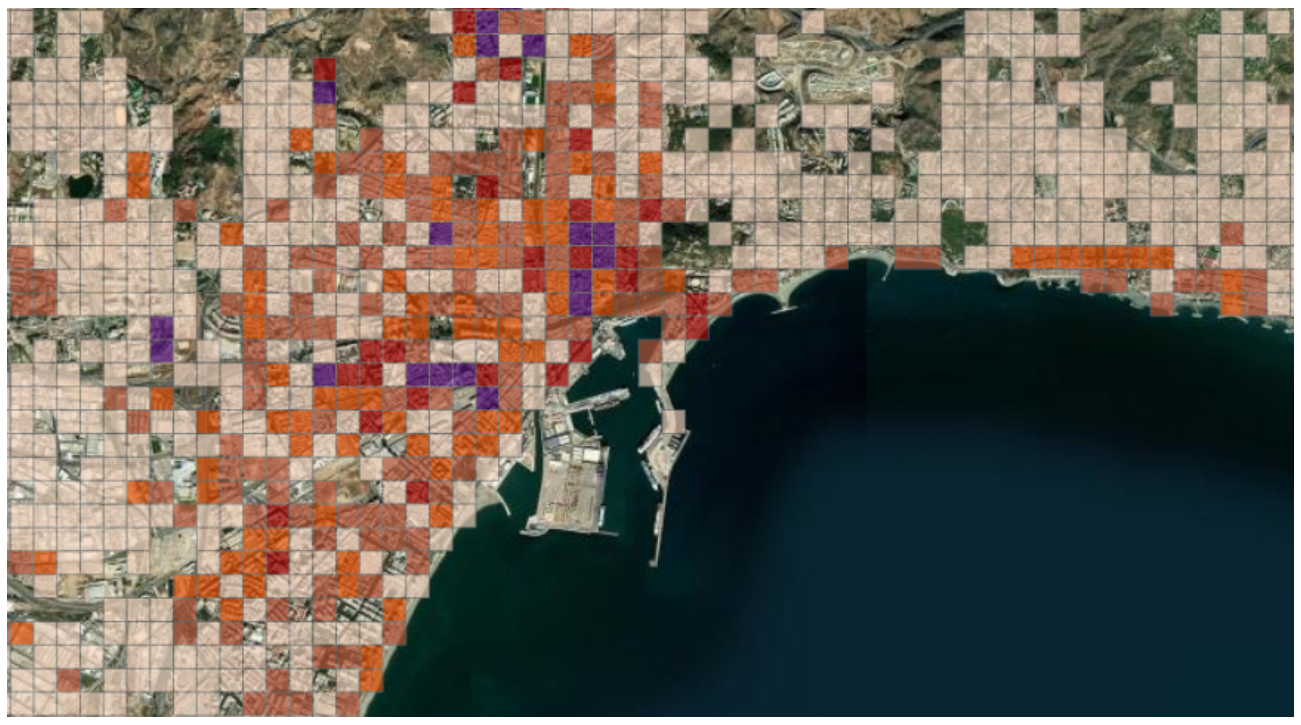

**Fig. 7.** Gridded urban environment (right).

*C. The agents*

In the model, the citizen agents, described by the subset C = {$c_1$, $c_2$, …, $c_n$}, represent the conventional actions performed by the persons who live in a city. Incorporating the basis of the RAT, these agents follow a pre-established daily routine throughout. There are four actions these agents will be able to perform throughout each day, which are listed next:

- *Rest (R):* the citizen moves to the cell assigned as home, which has been designated randomly in terms of the actual population of the districts.
- *Work (W)*: the citizen will go to the cell designated as work, randomly assigned among the habitable cells. Also note that the agent in this state could lose the job.
- *Leisure (L)*: the citizen will go to a cell chosen with the purpose of entertainment.
- *Looking for a job (J)*: the citizen will go to the cell assigned as home, that has been designated randomly following the actual population of the districts, and will have a probability of finding a job.

Three different roles of citizen agents can be identified in terms of their daily habits: A *morning citizen*, $c_i^M$, represents a person who either works, studies, is looking for job, or does not work for being a child or a retired person. The sequence of actions usually performed by this agent in a day is: $A(c_i^M) = (W \cup J, L, R)$. An *afternoon citizen*, $c_i^A$, who works in the afternoons and has a set of daily actions of $A(c_i^A) = (L, W, R)$; and a *night citizen*, $c_i^N$, who works at night, with the set of actions $A(c_i^N) = (R, L, W)$. The police unit agent ($p_i$) represents the human resources of the LEAs deployed in the model, which could be police patrols, couple of officers, etc. According to deterrence theory [52], police units have a very high deterrent effect when they are present in an area. Therefore, each police unit in the model will act as a crime reducer in the areas where it is deployed, thus representing changes in crime risk in terms of police presence in the area.

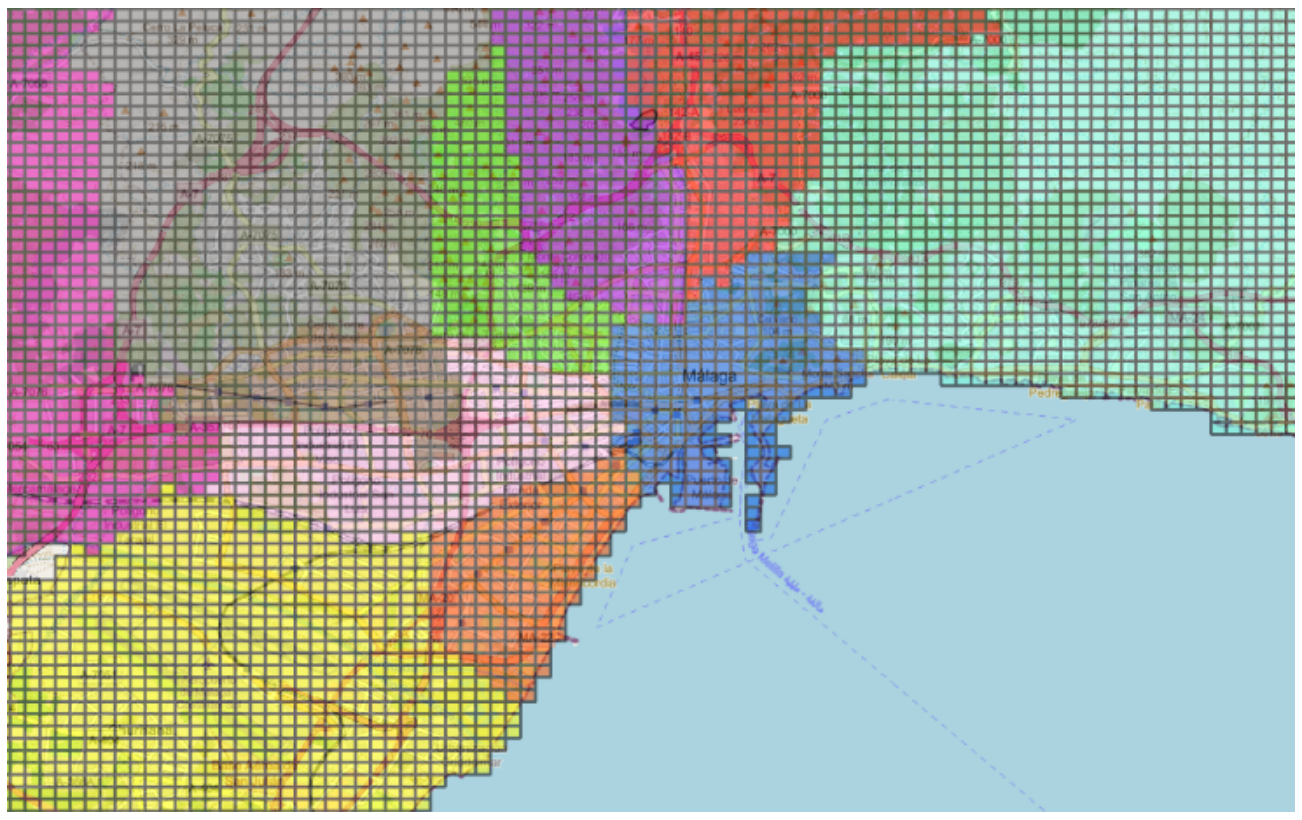

**Fig. 8.** Distribution of districts in the gridded urban environment. The color of the cell identifies a district of the city

Table IV summarizes the attributes of the simulation agents, citizens, and police units, and the urban environment cells. These attributes are used during the simulation to maintain the state of each component. The urban cells are described by their coordinates on the grid. Furthermore, each cell stores information such as the name of its district, whether it contains buildings, or if it is suitable for walking. Two attributes are related to the criminal behavior in that zone: the criminal power (that will be explained later) and the number of crimes committed there during the simulation. Citizen agents have attributes indicating their role, i.e., morning, afternoon, and night citizens, their home and work cell, and their employment status. Regarding the police unit agent, it has three attributes: the cell where it is deployed, its current cell in the simulation, and its time shift.

### D. Model global parameters

Our crime model requires some global parameters whose values need to be defined before running any simulation. These parameters provide the model with the necessary flexibility to represent different urban environments. The values of some parameters could be obtained directly from statistics on, for instance, population density, rates of unemployment, etc. Also, there are other input parameters, such as the number of virtual citizens and police units to be modeled in the simulation, which are computed from these input values but in terms of equivalences with real data. The *citizen_agent_rate* parameter, for example, computes the equivalence between an agent and a real citizen, giving us a measure of how many citizens each agent represents. Finally, the rest of parameters need to be calibrated from historical data on crime. Table V shows all the input model global parameters organized into two categories: those related to social and demographic issues and parameters involving crime simulation aspects. Besides the name of the parameter, Table V also contains the domain to which it belongs, its symbol (if applicable), its description, and how its value is obtained. As can be seen in the last column, the parameter value can be either provided as a model input parameter, obtained from historical data, computed from other parameters, or in need of being calibrated.

TABLE IV
ATTRIBUTES OF THE URBAN MODEL COMPONENTS

| Comp. | Attribute | Description | Data type |
|---|---|---|---|
| Cell | *x* | Coordinate *x* on the grid | integer |
| | *y* | Coordinate *y* on the grid | integer |
| | district | Name of the district where it is located | string |
| | habitable | Are there buildings registered in this part of the city? | Boolean |
| | walkable | If it is not sea or part of another municipality. | Boolean |
| | criminal_power | Computed value of increase in crime | double |
| | crime_counter | Number of crimes committed | integer |
| Citizen | home_cell | Location of home | cell |
| | work_cell | Location of workplace | cell |
| | role | Type of citizen in terms of the time slot where he/she works | enum{*morning*, *afternoon*, *night*} |
| | unemployed | Is he/she unemployed? | Boolean |
| Police Unit | current_cell | Cell where it is currently located | cell |
| | shift | Time slot where it patrols | enum{*morning*, *afternoon*, *night*} |
| | work_cell | Location of workplace | cell |

## VI. THE MODEL SIMULATION

Once the simulation environment has been described by a grid of cells, the set of agents that will participate in the simulations has been defined, and the global parameters that characterize the simulations have been established, this section is devoted to explaining the behavior of the crime simulation model. Firstly, and to contribute to a better understanding of the behavioral algorithm, we explain the theoretical model of our agent-based approach by introducing some concepts related to the criminal behavior, such as the criminal power of a cell or the probability of a criminal offense being committed by a citizen. Other relevant simulation issue is how citizens organize their leisure time. Finally, the main simulation algorithm is described, which basically expresses how we have modeled the daily behavior of the citizens in the urban environments. For all these purposes, we have incorporated in our model the expert knowledge and experience of fields practitioners and criminological theories.

### A. The criminal power indicator

To make a realistic representation of the crimes in an urban environment, we need to measure somehow the degree of crime incidence in each part of that environment. Crime simulation strongly depends on the number of offenses committed in each part of the city. Accordingly, for each cell of the urban environment, we first grouped and summed the crimes committed per year and cell of the simulation grid. Then, the number of crimes in the last year available in the input crime dataset (2018 in the case of Málaga) was predicted through a polynomial regression per cell, approximated with data from the rest of the years (2010 to 2017 in the case of Málaga). Once this estimation was obtained, this value was converted into a "multiplier", called *criminal power* of the cell, that was used in the simulation. The *criminal power* $\in (0, \infty)$ makes it possible to calculate how frequent it was to commit a crime in that cell with respect to the rest of the grid. Equation (1 shows how the criminal power value of a cell $e_{x,y}$, placed at coordinates *x* and *y*, can be computed:

$$\text{criminal_power}(e_{x,y}) = \frac{\max\left[\text{prediction}(e_{x,y}) + \alpha(e_{x,y}) * \left|\text{prediction_error}(e_{x,y})\right|, 1\right]}{\dfrac{\sum_{\substack{0\leq i<m \\ 0\leq j<n}} \text{prediction}(e_{i,j}) + \alpha(e_{i,j}) * \left|\text{prediction_error}(e_{i,j})\right|}{m * n}} \qquad (1)$$

where $\texttt{prediction}\,(e_{x,y})$ is the number of crimes estimated for the simulation year in the corresponding cell, $\textit{prediction_error}\,(e_{x,y})$ the error made in the polynomial regression approximation of that cell, and $\alpha(e_{x,y})$ a pseudorandom number generation function that takes values between –1 and 1, generated for the cell $e_{x,y}$. Finally, $m$ and $n$ are the width and height of the grid. In order to provide each cell with a minimum probability for crimes to occur and to avoid negative numbers, the number of crimes in the cell was calculated as the maximum between one and the result of the prediction of crimes in the corresponding cell for the year to be predicted in the simulation, plus a random value computed from the possible error made multiplied by a random coefficient generated from the function $\alpha$ and, finally, divided by the mean of all predictions plus the random value for that prediction.

### B. The probability of committing a crime

The factors that could lead a person to commit a crime are based on his/her employment status (*unemployed* attribute), the number of citizens in the cell where he/she currently is, the cell offense's attractiveness of the cell (`criminal_power`), and the model parameters involving the probability of making an offense in terms of the employment status. Equation 2 computes the probability, $P_{crime}(c_i, t_j)$, that a certain citizen $c_i$ will commit a crime in the cell where he/she is located at time $t_j$.

$$P_{crime}(c_i, t_j) = \frac{criminal_power(where(c_i, t_j))\ r_c\ \varepsilon\left(where(c_i, t_j), t_j\right)\ \left[r_u^{u(c_i,t_j)}\ r_e^{1-u(c_i,t_j)}\right]}{z\left(where(c_i, t_j)\right)} \quad (2)$$

In Equation 2, $z$: C × T → ℜ is a function relating each cell with the number of citizens placed there, *where*: {C ∪ P} × T → {E ∪ ∅} is a function mapping the agent with their current location; $r_c$ the *citizen_agent_rate*; $r_u$ and $r_e$ the unemployment and employment offense rates, respectively; and $u$: C × T → {0,1} a function indicating whether or not a certain citizen is currently unemployed at a certain time. The function $\varepsilon$: C × T → ℜ calculates the effect of the presence (or absence) of police units in the corresponding cell, using the number of police units in the last month and two simulation input parameters: *increase_no_police* ($\psi$) and *police_reduction* ($\rho$), where the first parameter determines the maximum increase in crime in a cell when no police unit is visiting (reaching its maximum value after 30 days of no visits), and the second parameter determines the crime reduction effect of a police unit when it is deployed in a cell. As shown in Equation 3, the function returns 1 if there are no police agents deployed in the simulation. The *police* function (Equation 4) measures the number of police units deployed in a cell over the last 30 days and *is_in* (Equation 5) is a Boolean function that returns 1 if a police unit is in a given cell at a given time.

$$\varepsilon(e_{x,y}, t_j) = \begin{cases} 1, & \forall p_i \in P, where(p_i, t_j) = \emptyset \\ 1 - \dfrac{police(e_{x,y}, t_j)}{30}(\psi + \rho) - \rho, & \nexists p_i \in P, where(p_i, t_j) = e_{x,y} \\ 1 - \rho, & otherwise \end{cases} \quad (3)$$

$$police(e_{x,y}, t_j) = \sum_{\forall p_i \in P} \sum_{k=0}^{30} is_in(p_i, e_{x,y}, t_j - k) \quad (4)$$

$$is_in(p_i, e_{x,y}, t_j) = \begin{cases} 1, & where(p_i, t_j) = e_{x,y} \\ 0, & otherwise \end{cases} \quad (5)$$

*C. The leisure of the citizen agents*

Beside working and resting at home, leisure time takes one third of the day of the citizen agent. We have modeled the destination of each citizen during his/her leisure time through a *leisure* function. Two hypotheses have been considered: people regularly enjoy their spare time in a place close to their residence, and secondly, people sometimes travel to the city center to spend their time there. Consequently, the function uses the model parameters `nearby_leisure_probability` and `downtown_leisure_probability` to determine the cell where the citizen will spend his/her leisure time. In the case of having his/her leisure time near home, the destination cell is selected randomly in a radius of 5 cells around the home cell. The function $leisure(c_i, t_j, l_k): C \times T \times L \rightarrow E$, is a pseudorandom number generation function that sets the citizen's leisure destination at a certain time in terms of a leisure scope, $L = \{l_N, l_D, l_H\}$, i.e., near home ($l_N$), at downtown ($l_D$), or at home ($l_H$). The *leisure* function fulfills the following conditions (Equation 6) regarding the values of the *scope*.

*D. Behavior of the simulation model*

The main behavior of the simulation is detailed in Algorithm 1. As can be seen, the simulation tries to mimic the behavior of the city during a year. Each iteration corresponds to a day, which in turn can be divided into the three time slots that are relevant from the perspective of police unit deployment strategies, that is, morning, afternoon, and night. In each time slot, the simulation performs in the same way. The behavior of each citizen agent, i.e., the sequence of actions he/she performs, depends on the citizen agent role and is also modeled through a set of probabilistic rules. For each citizen in the virtual world, his/her daily behavior is simulated, which is mainly conditioned by three factors: whether he/she is employed or not and, if so, what his/her work shift and the time slot of the day are. Thus, if the citizen has a work in the corresponding time slot, a random probability value is generated, and if this value is under the `lose_job_probability` threshold, the citizen loses his/her job. In this line, if the citizen is unemployed, he/she will find a job whenever if a random probability value is under the `find_job_probability` threshold. Otherwise, the behavior of the citizen will consist in moving to a leisure place or resting in terms of his/her sequence of actions described in subsection IV.C. Accordingly, in the case of having leisure time, it could take place near home or downtown, as explained in subsection VI.C. Finally, in the scope of this time slot, a citizen could try to commit a crime in terms of the function described in section VI.B. Once a crime is committed, the offenses counter of the corresponding cell is increased.

TABLE V

MODEL GLOBAL PARAMETERS

| Parameter | Domain | Meaning | Value |
|---|---|---|---|
| ***Social and demographic parameters*** | | | |
| Total population | **N** | Real population of the urban environment. In the case of Málaga, the current value is **572,260** citizens | Obtained from data |
| District population | **N** | Population of each district (see Table 2) | Obtained from data |
| Number of citizens (agents) | **N > 0** | **1,000** citizens | Input parameter |
| Number of police units (agents) | **N > 0** | Police units to be deployed | Input parameter |
| *citizen_agent_rate* ($r_c$) | **R+** | **572.26** (total population/number citizens) | Computed |
| *find_job_probability* | **[0, 1]** | **0.005** (calibrated to be around 30% unemployment during the simulation) Likelihood of finding a job | Calibrated |
| *lose_job_probability* | **[0, 1]** | **0.0022** (calibrated to be around 30% unemployment during the simulation) Likelihood of losing the job | Calibrated |
| *nearby_leisure_probability* | **[0, 1]** | Likelihood of spending leisure time close to home within a 1 km radius (i.e., 5 cells) | Calibrated |
| *downtown_leisure_probability* | **[0, 1]** | Likelihood of spending leisure time in downtown | Calibrated |
| ***Simulation crime parameters*** | | | |
| Total offenses | **N** | **304,125** geolocated offenses in Málaga | Obtained from data |
| *offense_rate* | **R+** | **5.39 * $10^{-5}$** (calculated by dividing the total number of offenses by the time the perpetrators have had to offend, 8 years [2010–2017]) | Computed |
| *unemployment_related_increase_in_crime* ($\mu$) | **[0, 1]** | Rate of increase in the probability of committing a crime because of being unemployed | Calibrated |
| *unemployed_offense_rate* ($r_u$) | **R+** | Likelihood that a crime is committed by an unemployed person; this parameter takes into account the *offense_rate* and the unemployment-related increase in crime | Computed |
| *employed_offense_rate* ($r_e$) | **R+** | Likelihood that a crime is committed by a person with a job; this parameter also considers the *offense_rate* and the unemployment-related increase in crime | Computed |
| ***Police deployment parameters*** | | | |
| *police_reduction* ($\rho$) | **[0, 1]** | Crime prevention of police units | Input parameter |
| *increase_no_police* ($\psi$) | **[0, 1]** | Increase in crime due to lack of police units | Input parameter |

As can be seen in Algorithm 1, every two weeks, the crime probabilities in the cells are updated, according to Equation 1, and the police unit deployment strategy could also be upgraded using the `redistribute_police_units()` procedure. Furthermore, when police unit agents are deployed, each day and in each time slot, i.e., morning, afternoon, and night, the presence of police units in all cells is reset, and every morning the `police_influence` in each cell is decreased to achieve a realistic representation of the security of an area. The idea is simple: if an area is always under surveillance by police officers, offenders will try to avoid it. But if the police officers abandon it, the offenders are likely to return. Following the Deterrence Theory and the RAT, if an area is insecure due to the absence of any guarantor or guardian to ensure safety, criminals will have a greater incentive to commit crimes. All the movements due to policing strategies lead to an increase of the number of times some patrol has visited the destination cell (i.e., `police_influence` attribute) and also the presence of the police in that cell (`police_here` attribute) is recorded. Finally, the function modeling the possibility of committing a crime has also been updated to consider the influence of the police unit presence in a cell as a disincentive to crimes (see Equation 2). This function takes into account the following parameters: *i)* the employment situation of the citizen; *ii)* the general crime rate of the city (*unemployed_offense_rate* global parameter if the citizen is unemployed and *employed_offense_rate* otherwise); *iii)* the crime attractor of the cell (`criminal_power` function) where he/she is located; i*v)* whether police units are present (`police_here` attribute); and *v)* if police units have been regularly in that cell (`police_influence` attribute, which designates the number of times a police unit has been in that cell in the last month).

## VII. MODEL CALIBRATION AND VALIDATION

Many studies have remarked the problem of calibrating ABMS-based systems [53] [54]. As mentioned before, some global input parameters of our model (Table V) that are not observable need to be calibrated. In this case, the goal of model calibration is precisely to infer the values of those calibrated parameters in such a way they will fit our model in terms of the record of crimes committed in the city, leading thus to a suitable DS of the urban environment. To perform such model calibration, crimes in the period comprising all years available except the last one (2010–2017 for Málaga) are used as inputs of the model. The results obtained are tested with the crimes committed in the last year (2018). Note also that during the calibration process no police unit agents are deployed in the model. All information regarding crime incidence is obtained directly from historical data, and therefore, criminal power is calculated solely using this information.

**Algorithm 1** Simulation of the Urban Crime

```
day = 0
    generate citizens
    while day < 365 do // One year, 365 days
       if day%14 == 0 then // Every 2 weeks (14 days)
          update_cell_criminal_power_probabilities()  // criminal_power(x, y)
          redistribute_police_units()  // Patrol assignment algorithm
       end-if
       foreach time_day in {morning, afternoon, night} do
          deploy police units
          foreach citizen in all_citizen do
                citizen performs the next corresponding action
                switch citizen current action do
                      case 'work':
                         destination = citizen workplace
                         if citizen randomly loses his/her work then
                             citizen changes his/her state to unemployed
                         end-if
                      case 'find job':
                         if citizen randomly finds a job then
                             citizen changes his/her state to employed
                         end-if
                      case 'rest':
                         destination = citizen home place
                      case 'leisure':
                         select randomly a leisure scope, L = {l_N, l_D, l_H},
                         destination = compute leisure function value
                end-switch
                citizen moves to destination
                prob_crime = compute citizen's probability of committing a crime
                if generate random value < prob_crime then
                      citizen commits a crime
                end-if
          end-foreach
       end-foreach
       day++
    end-while
```

In the case of Málaga, the three following global parameters were calibrated: *unemployment_related_increase_in_crime*, which directly affects the crime rates, and the probabilities of *nearby_leisure_probability, downtown_leisure_probability*, which determine how the agents move during the simulation. After an exploratory process of model simulations using different combinations of parameter values, the eight best configurations that were tested during this calibration process to fit the city's crime behavior model with data of 2018 are described in Table VI. Each configuration of values was run at least 200 times to avoid stochastic situations produced by chance and to draw reliable averages of the model behavior with these input parameters. The number of citizens deployed in each simulation was 1,000 agents (each one representing approximately a group of 517 people since there are around 517,000 people registered in Málaga city). The main reason for choosing this value was that this number of agents covered all the cells in the city and did not consume an excessive amount of time when carrying out the simulations.

TABLE VI

PARAMETER VALUES FOR MODEL CALIBRATION

| **Config. ID** | ***unemployment_ related_increase_ in_crime*** | ***nearby_ leisure_ probability*** | ***downtown_ leisure_ probability*** |
|---|---|---|---|
| 1 | 0.10 | 0.50 | 0.075 |
| 2 | 0.10 | 0.50 | 0.100 |
| 3 | 0.10 | 0.60 | 0.075 |
| 4 | 0.10 | 0.60 | 0.100 |
| 5 | 0.15 | 0.50 | 0.075 |
| 6 | 0.15 | 0.50 | 0.100 |
| 7 | 0.15 | 0.60 | 0.075 |
| 8 | 0.15 | 0.60 | 0.100 |

Validating a complex model is not straightforward, especially if the model involves the definition and configuration of many parameters [55]. To evaluate the performance of the results, the following metrics, widely used in the field of predictive policing, are used: *predictive accuracy index* (PAI), *predictive efficiency index* (PEI) and *forecast accuracy index* (FAI). PAI is a metric that interprets the density of crime in the predicted hotspots regarding the total crime in the region under study [56]; the best value of PAI is often called PAI*. PEI is the calculation of how good the prediction is with respect to PAI* [57]. Lastly, FAI measures the proportion of crimes falling in the predicted hotspots normalized by the total number of crimes. With these indicators, we can explore whether the model can identify the most conflictive areas accurately.

$$PAI = \frac{hit\ rate}{area\ coverage} \quad (7)$$

$$FAI = \frac{(n^{real}/n^{simulation})}{(N^{real}/N^{simulation})} \quad (8)$$

Equation 7 describes how PAI is computed: *hit rate* is the percentage of predicted crimes within the corresponding area or defined hotspots area, and *area coverage* is the prediction area in relation to the whole study area. The value of PAI is 1 if the model predicts all the crimes in the entire study area, while it can be greater than 1 if the selected coverage area is small. The area coverage can be understood as the area the police can cover. Typical values for coverage area range from 3% to 20% [46] [56] [58]. Equation 8 describes FAI [26]; this metric represents the difference between simulated and actual crimes within the hotspots relative to the entire city, $n^{real}$ represents the number of real crimes in the predicted hotspots and $N^{real}$ the number of crimes in the whole city. A simulation run with FAI closer to 1.0 is considered to replicate more accurately the relative sizes of spatial clusters in real crimes. Metrics of supervised classification such as precision, recall, or F-measure are also considered. *F-measure*, i.e., the harmonic mean between precision and recall, allows synthetizing both indicators in the same metrics and thus provides a better comparison of the results among different configurations. It should be mentioned that different measurements have been made for all these metrics by changing the number of crimes for a cell to be considered a hotspot and therefore a hit; specifically, measurements have been made with at least 1, 10, 50, 100, and 200 crimes.

TABLE VII
PAI, PEI AND FAI RESULTS OF THE DIFFERENT CONFIGURATIONS.

| ID | PAI | | | | PEI (PAI / PAI*) | | | | FAI | | | |
|---|---|---|---|---|---|---|---|---|---|---|---|---|
| | 3% | 5% | 10% | 20% | 3% | 5% | 10% | 20% | 3% | 5% | 10% | 20% |
| 1 | 12.41 | 9.7 | 6.76 | **4.25** | 0.932 | 0.932 | 0.953 | **0.966** | **0.816** | **0.814** | **0.856** | **0.922** |
| 2 | **12.45** | 9.72 | 6.78 | **4.25** | **0.935** | 0.934 | 0.956 | **0.966** | 0.808 | 0.807 | 0.854 | 0.918 |
| *3* | 12.24 | 9.69 | 6.75 | **4.25** | 0.919 | 0.931 | 0.952 | **0.966** | 0.804 | 0.812 | 0.85 | 0.918 |
| *4* | 12.41 | **9.78** | 6.76 | **4.25** | 0.932 | **0.939** | 0.953 | **0.966** | 0.808 | 0.804 | 0.847 | 0.916 |
| *5* | 12.37 | 9.71 | 6.77 | **4.25** | 0.929 | 0.933 | 0.955 | **0.966** | 0.813 | 0.813 | 0.854 | 0.919 |
| *6* | 12.26 | 9.69 | **6.79** | 4.24 | 0.92 | 0.931 | **0.958** | 0.964 | 0.794 | 0.802 | 0.852 | 0.916 |
| *7* | 12.18 | 9.7 | 6.77 | **4.25** | 0.914 | 0.932 | 0.955 | **0.966** | 0.796 | 0.809 | 0.85 | 0.916 |
| *8* | 12.25 | 9.74 | 6.75 | 4.24 | 0.92 | 0.936 | 0.952 | 0.964 | 0.792 | 0.805 | 0.844 | 0.913 |
| * | *13.32* | *10.41* | *7.09* | *4.4* | *1* | *1* | *1* | *1* | *0.933* | *0.923* | *0.928* | *0.975* |

As shown in Table VII and Figs. 9 mad 10, the PAI, PEI, and FAI results for coverage area range from 3% to 20%. Last row of Table VII displays the results obtained using the locations with the highest number of crimes in the test dataset, indicating the optimal results attainable within this environment. Table VIII and Fig. 11 illustrate the precision, recall, and F-measure in terms of 1 to 200 offenses per cell.

We also looked at the differences between the crimes committed in each district of the city, according to each of the configurations. The aim was to compare the crimes that occurred in 2018 with those predicted by our model in each district in each configuration. To do this, we ran the simulation of each configuration 100 times and thus obtained the average results of the crimes per district.

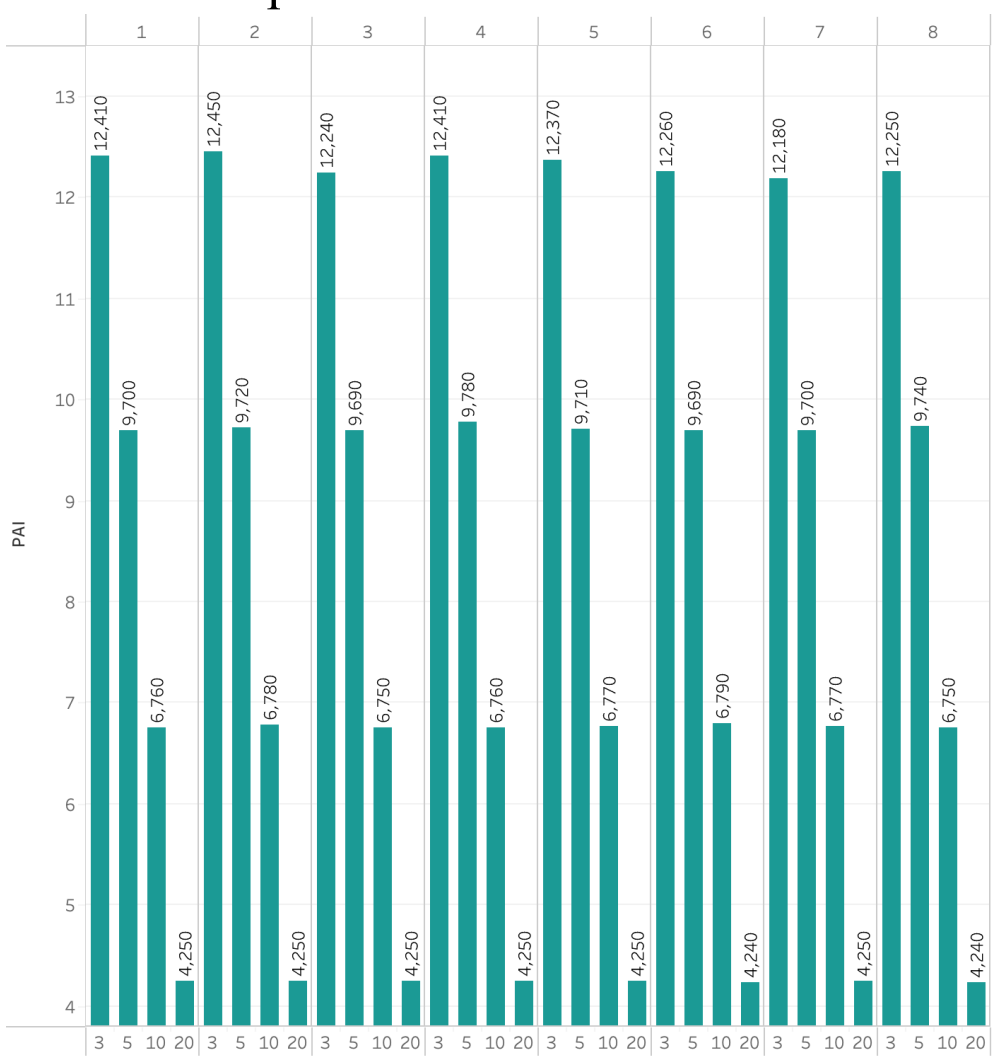

**Fig. 9.** PAI results of the different configurations

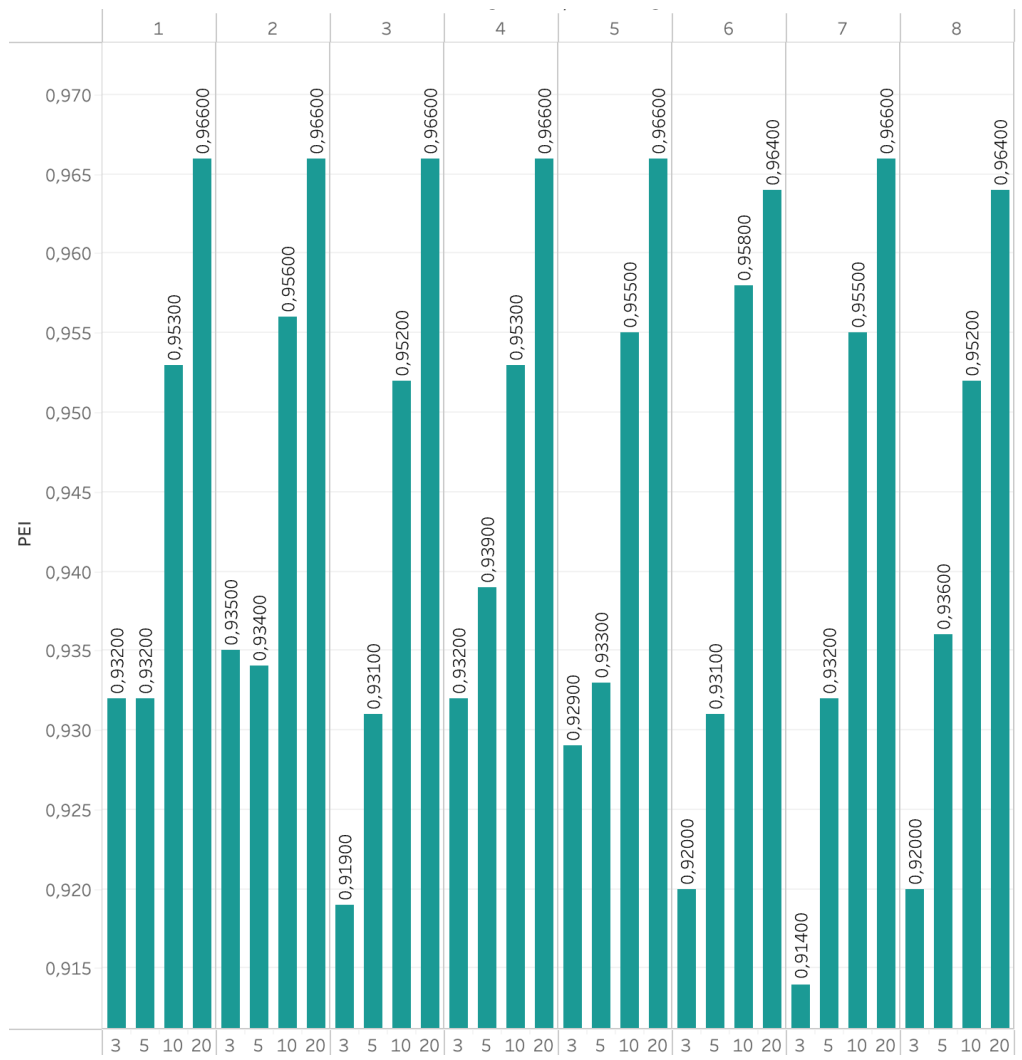

**Fig. 10.** FAI results of the different configurations

With these results, we performed some statistical analysis to assess analyze whether or not the differences between the crime rates in each district were significant enough. The paired Student's t-test is a parametric statistical procedure that allows the null hypothesis that the means of two populations are equal to be tested against the alternative hypothesis that they are not, taking each measure twice, resulting in a pair of observations. As can be seen in Table IX, the p-value (0.417) is much higher than the standard significance level of 0.05, so we cannot reject the null hypothesis. We also calculated another statistical test analogous to the paired t-test, the Wilcoxon test, which is used when sample means are not of interest. The null hypothesis for this test is also analogous, and the results again show that the null hypothesis cannot be rejected. Finally, we calculated the Spearman's rank correlation coefficient between the real data of 2018 (Table X and Fig. 12) and the values provided by our simulation model, obtaining a value greater than 0.84 ($p \leq .002$) in all cases, which indicates a significant and positive relationship between both variables. Note that Spearman's rank is the most appropriate correlation coefficient when the number of observations is small.

TABLE VIII

PRECISION, RECALL AND F-MEASURE IN TERMS OF THE NUMBER OF CRIMES IN A CELL TO BE CONSIDERED A HOTSPOT (HIT)

| | ***Crimes per cell to be considered a hotspot*** | | | | | | | | | | | | | | |
|---|---|---|---|---|---|---|---|---|---|---|---|---|---|---|---|
| | 1 | | | 10 | | | 50 | | | 100 | | | 200 | | |
| ***ID*** | *Prec.* | *Rec.* | *F-m.* | *Prec.* | *Rec.* | *F-m.* | *Prec.* | *Rec.* | *F-m.* | *Prec.* | *Rec.* | *F-m.* | *Prec.* | *Rec.* | *F-m.* |
| 1 | 0.90 | **0.83** | **0.87** | **0.89** | 0.76 | **0.82** | 0.69 | **0.90** | **0.78** | **0.47** | **0.88** | 0.61 | **0.59** | **0.83** | **0.69** |
| 2 | 0.90 | **0.83** | **0.87** | **0.89** | 0.76 | **0.82** | 0.69 | 0.88 | 0.77 | 0.45 | **0.88** | 0.60 | 0.56 | **0.83** | 0.67 |
| 3 | **0.91** | **0.83** | 0.86 | 0.88 | **0.77** | **0.82** | 0.68 | 0.89 | 0.77 | 0.44 | 0.86 | 0.58 | 0.53 | **0.83** | 0.65 |
| 4 | **0.91** | 0.82 | 0.86 | 0.88 | **0.77** | **0.82** | 0.68 | **0.90** | 0.77 | 0.44 | **0.88** | 0.58 | 0.53 | **0.83** | 0.65 |
| 5 | **0.91** | 0.82 | 0.86 | **0.89** | 0.76 | **0.82** | **0.71** | 0.87 | **0.78** | **0.47** | 0.87 | **0.62** | **0.59** | **0.83** | **0.69** |
| 6 | **0.91** | 0.82 | 0.86 | **0.89** | 0.76 | **0.82** | 0.70 | 0.88 | **0.78** | **0.47** | **0.88** | 0.61 | **0.59** | **0.83** | **0.69** |
| 7 | 0.90 | 0.82 | 0.86 | **0.89** | **0.77** | **0.82** | 0.68 | 0.89 | 0.77 | 0.45 | **0.88** | 0.60 | 0.53 | **0.83** | 0.65 |
| 8 | 0.90 | 0.82 | 0.86 | 0.88 | **0.77** | **0.82** | 0.68 | 0.88 | 0.77 | 0.43 | **0.88** | 0.58 | **0.59** | **0.83** | **0.69** |

In the end, configurations 1, 5, and 6 yielded the best results, with all three achieving very similar outcomes. Ultimately, configuration 1 was chosen because it stands out in FAI ahead of all of them, while it either matches or surpasses the others in the remaining metrics. Additionally, this configuration generates a more upward pattern in the number of crimes committed, which can be advantageous when the objective is to explore police or socio-economic strategies to reduce crime. Fig. 13 shows the results of configuration 1 on the city map. This figure contains two heatmaps of the city of Málaga: the one on the left illustrates with real data the crimes committed in 2018, while the one on the right shows the heatmap resulting from the simulation of that year.

It should also be mentioned that there are two areas belonging to the southern part of the city, in the district of Churriana, i.e., the airport and Plaza Mayor (a large shopping mall), which are not well replicated due to the difference in the flow of people with respect to the rest of the city. Due to their location (people move there to travel or go shopping), they have much more people flow than they would otherwise. Since there are no people living near the airport, the flow of people in these areas is not well represented, and, therefore, the crimes committed are inadequately reproduced in the model. This problem is not relevant in our case, since our objective is to model the urban environment of the city, and in consequence, it does not make much sense to incorporate these areas into our model. Therefore, and except for the differences that have been discarded in our model, in the comparison between the heatmap obtained with the historical crime data of 2018 and the heatmap generated with the simulation data, a coincidence of the general pattern of crime in the city is clearly recognized.

## VIII. DISCUSSION

The grid representation for modeling the urban environment proves to be a good solution due to the lack of precise geolocation of crime in our data. First, it eliminates the need to work with street layout geometry, which in the case of European cities is often quite complex. Secondly, as mentioned above, it allows us to group crimes into zones, separating real hotspots from others fictitiously generated, due to non-crime-related factors such as the existence of large avenues (with lengths of several kilometers). Crimes recorded by police officers are often not correctly geolocated, indicating only the name of the place (street, avenue, road, etc.) where the crime occurred and thus giving rise to these false hotspots in large ubications. Moreover, the size of our gridded cell (192 m side cells) is smaller, and thus more precise, than others previously applied in empirical criminological studies [56] or at least similar [48] [49] [59]. The predictive values of our model are better than (or similar to) those of previous works, even compared to others performed on street segments [38], as we will elaborate quantitatively below.

Since in many datasets, the time of the crime event is imprecise, missing, or unknown, and in order to simplify the model, each day is divided into three periods of equal length: morning, afternoon, and night. This distinction coincides with police patrol shifts and avoids the need to apply a pathfinding algorithm so that agents have to move from one point to another in the city. Some studies, however, have shown that time can have an important impact on predicting performance [49] [60]. Consequently, our model could be improved by adding crime rates according to the part of the day. This would allow testing the distribution of police units in different shifts and could be useful to test the variability of risk during the day.

TABLE IX

PAIRED T-TEST AND WILCOXON SIGNED-RANK TEST RESULTS

| **Configuration** | **Test** | **Statistic** | **z** | **df** | **p** |
|---|---|---|---|---|---|
| All | Student | 0.847 | | 10 | 0.417 |
| | Wilcoxon | 34.000 | 0.089 | | 0.966 |
| 1 | Student | -0.819 | | 10 | 0.432 |
| | Wilcoxon | 32.000 | -0.089 | | 0.966 |
| 2 | Student | -0.832 | | 10 | 0.425 |
| | Wilcoxon | 32.000 | -0.089 | | 0.966 |
| 3 | Student | -0.919 | | 10 | 0.380 |
| | Wilcoxon | 32.000 | -0.089 | | 0.966 |
| 4 | Student | -0.942 | | 10 | 0.368 |
| | Wilcoxon | 32.000 | -0.089 | | 0.966 |
| 5 | Student | -0.716 | | 10 | 0.491 |
| | Wilcoxon | 34.000 | 0.089 | | 0.966 |
| 6 | Student | -0.767 | | 10 | 0.461 |
| | Wilcoxon | 35.000 | 0.178 | | 0.898 |
| 7 | Student | -0.857 | | 10 | 0.411 |
| | Wilcoxon | 32.000 | -0.089 | | 0.966 |
| 8 | Student | -0.882 | | 10 | 0.398 |
| | Wilcoxon | 32.000 | -0.089 | | 0.966 |

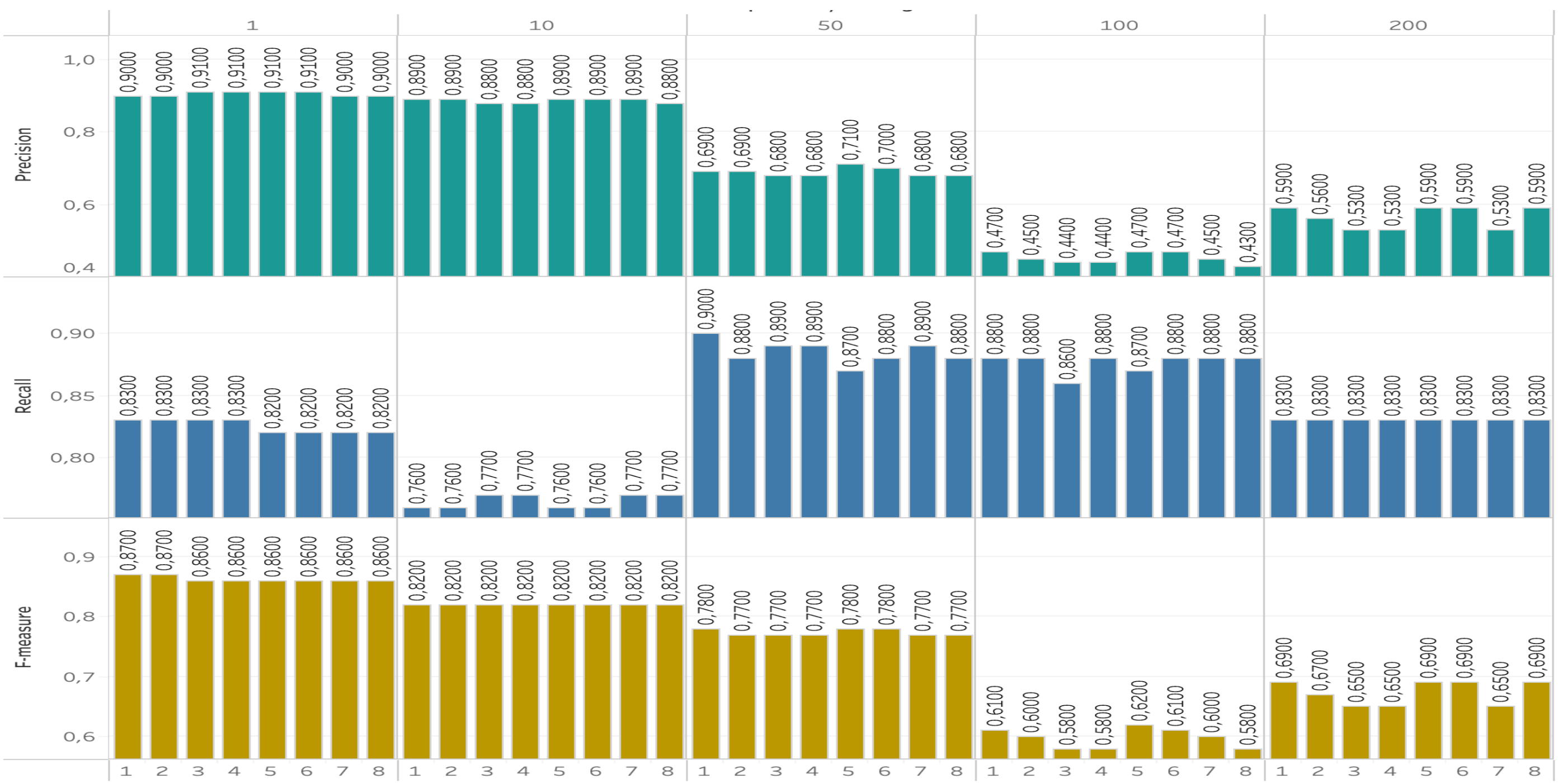

**Fig. 11.** Precision, recall, and F-measure results in terms of the number of crimes and configurations

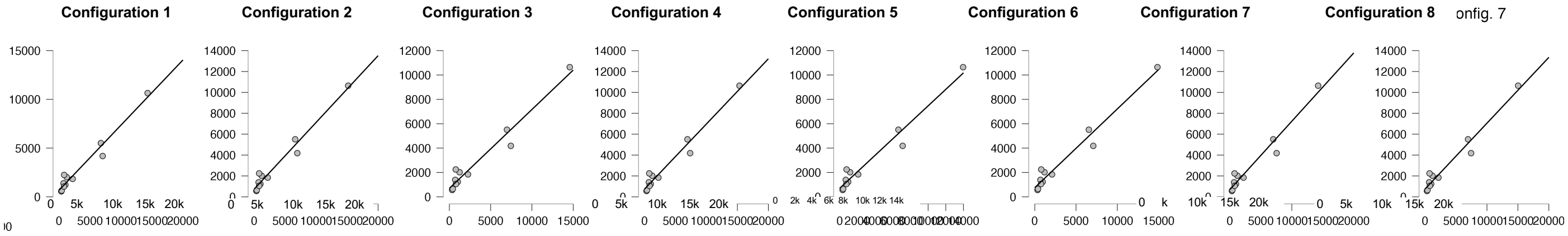

**Fig. 12.** Graphical plot of the Spearman's rank correlations between the real data and the output of the simulation model.

Crimes are complex human events based on multiple social factors that cannot be predicted with enough accuracy. For this reason, when criminals, citizens, and potential victims are modeled, different theoretical approaches can be followed in terms of the modelers' understanding of the crime theories. In our work, we model the citizen agents using RAT and CPT. RAT assumes that potential offenders and victims frequent certain places at certain times as part of their daily routines. Our simulation model tries to mimic the coarse-grained dynamics of people's movements in cities, not the reasons behind the commission of a crime, hence the simplification in decision-making by citizen agents. Fine-grained decision-making does not make sense in a model such as the one we propose, since there is not enough detailed spatial information to make realistic decisions. To do so, it would be necessary to specify the concentration of leisure areas, information about the presence of tourists in each area of the city (as they are potential victims for the most common type of crime, theft), etc. Moreover, our model does not distinguish between common citizen and offender agents. Like [25] and [26], we try to minimize the bias on crime opportunity locations from unconstrained target searching behaviors. RAT also suggests that motivated offenders are not exempt from fulfilling their daily schedules of legitimate activities, work and rest included. In contrast, other works, e.g. [40], distinguish between both types of agents in order to reduce the simulation time and the computational cost of modeling the offending decision-making process. Other proposals even exclude citizen agents from the simulation environment because they are not the direct targets of certain offenses, e.g. burglaries [44] [61]. In our approach, the offending propensity for every citizen agent is modeled taking into consideration the following factors that could lead a citizen to commit a crime combined with the criminological theories that support them:

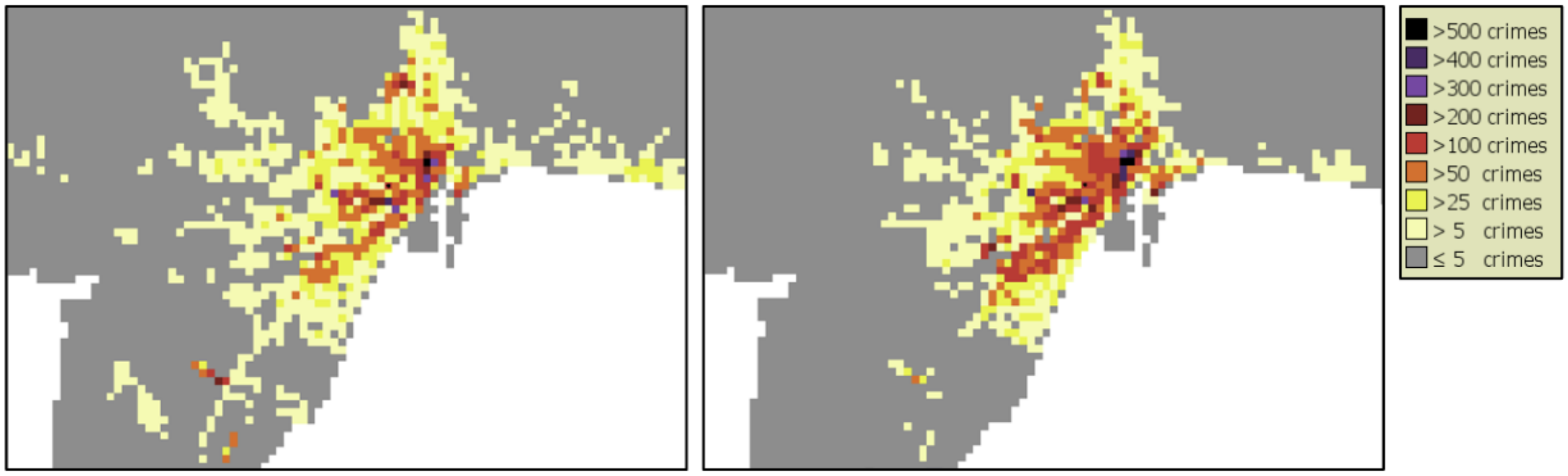

**Fig. 13.** Heatmap of the crimes committed in Málaga in 2018 (left) and map with the crimes that occurred in the simulation of the same year with configuration 1 (right).

TABLE X

SPEARMAN'S RANK CORRELATIONS BETWEEN THE REAL DATA AND THE OUTPUT OF THE SIMULATION MODEL

| **Config.** | **1** | **2** | **3** | **4** | **5** | **6** | **7** | **8** |
|---|---|---|---|---|---|---|---|---|
| **Spearman** | **0.981** | **0.845** | **0.845** | **0.845** | **0.891** | **0.981** | **0.845** | **0.845** |
| **p-value** | **< .001** | **.002** | **.002** | **.002** | **< .001** | **< .001** | **.002** | **.002** |

- The employment situation of the citizen: According to [62], employment status is often the primary feature to create the fundamental behavioral dissimilarities, as the activities of employed and unemployed people have different locations, starting times, and durations on weekdays. For that reason, our model uses different daily routines depending on the employment situation of the citizen agents. In addition, the employment status of the citizen agent and the parameters involving the unemployment rate of the city are also considered as factors to commit a crime. The use of this last rate is also aligned with the social disorganization theory (SDT) [63], which links crime levels to environmental attributes of the neighborhood such as the socio-economic status. Therefore, the employment status enables the creation of more precise behavior models for urban environments, where assuming uniform actions for all agents would be atypical [62]. This employment factor is a feature that differentiates our proposal from most existing models based on crime theories [28] [26] [40] [64].
- The criminal power of the cell where the agent is located: Following the theory of repeat victimization, which states that crime events are more likely to occur at locations at which other crime incidents have previously taken place, this function incorporates counts of previous crime incidents (based on historical data) into our predictive models and establishes how frequently a crime was committed in a cell.
- The presence and regularity of the police units in a cell, i.e., the number of times a police unit has been in that cell in the last month: deterrence theory states that crimes can be prevented when the costs of committing them are perceived by the offender to outweigh the benefits. General deterrence is the idea that the general population is dissuaded from committing a crime when it is observed that punishment necessarily follows the commission of a crime. In the same way, RCT states that offenders will weigh rewards against risks before committing the crimes.

In this work, we have also performed the calibration and validation of an instance of the DS platform with the information related to the city of Málaga. These two stages are often incomplete in many proposals of related literature, primarily due to the confidentiality surrounding crime data. In our study, various metrics such as PAI, PEI, FAI, F-measure, and other statistical analyses such as the Student's t-test or the Wilcoxon test have been explored. We truly believe that our validation procedure enables future researchers to compare their results with ours more easily. The predictive values achieved suggest that it is an accurate representation of the actual crime situation in the city: the values of PAI (3%) 12.41 and PAI (20%) 4.25 are higher than those of other approaches. For example, Adepeju et al. [58] exhibit a PAI (20%) of 2.99 for violence and 4.58 for shoplifting (using the Candem crime dataset) with a 250 m-side grid unit scale; Rummens et al. [46] only report a higher PAI value of 13.80 for street robbery, while reporting 3.77 for assaults in the bi-weekly predictions, but they do not specify the proportion of the area used, making the comparison not possible. Lin et al. [54] show a PAI (20%) of 3.39 using the 150 m-side square scale, and Roses et al. [42] report a PAI (3%) of 10.25 and a PAI (20%) of 4.51. In this last work, predictions are made on street segments rather than on grid cells, but our predictive values are even better or comparable to them. In [26], Zhu et al. get a PAI (7.2%) of 8.7295, which is lower than our PAI (5%) 9.7. Additionally, they achieve a higher result (1.09) in the FAI indicator, compared to our 0.814, but this value was observed in only one of their several simulations; in the rest, they attained a significantly low score. Regarding PEI, Lee et al. [57] [64] achieve a PEI score of 0.823 (82.3%), but they do not indicate the hotspot area. However, we outperform those results ranging from 3% to 20% of the city's surface, leading to an improvement from 0.932 (93.2%) to 0.966 (96.6%). Finally, the precision and recall results of our model, 0.47 and 0.88 respectively (see in Table VIII the fourth cell column group of over 100 crimes per cell), suggest that the model makes many false positives due to the bullish feature of our model. But, combining these results with those of 12.41 and 0.93 for PAI and PEI at 3%, it can be inferred that the model correctly selects which cells will have more crimes committed, even if it assigns more crimes to other cells in the model. Statistical tests such as Spearman's rank correlation, paired t-test, and Wilcoxon test also suggest that our model is able to properly simulate the criminal behavior of the urban environment.

Regarding the limitations, we are aware that modeling social environments is a challenge since it involves many latent factors that are difficult to be modeled. Data-driven systems strongly depend on the quality of their data. In our case, we have been able to instate our platform with a full dataset of crimes from a period of around 10 years, but this information is hardly available in any other city, since it must be provided by authorities. The architecture of the platform, however, considers the integration of multiple data sources that can be combined to enrich the information on each cell of the urban environment. We use grid cells for modeling spatial environment, so we cannot place crime where it occurs. Therefore, it would be desirable to combine a grid-based scheme with the inclusion of street networks that allow improving the agents' movement in urban environments. Anyway, and according to the literature, grid cells is a spatial modeling strategy suitable for most cities. Due to its macroscopic features, our model cannot predict crimes within a short time span slot, such as a day or a week. The factors involved in predicting within such a brief period require more complexity and encompass more considerations, such as atmospheric conditions or any festivities that may alter the behavior or mobility of LEAs. In the future, as can be seen in Section IX, we will study the creation and integration of multi-agent reinforcement learning algorithms to simulate and predict criminal behavior in our model in a smaller period of time. Another aspect our model lacks is the incorporation of the motivation to commit a crime. We have followed a macroscopic modeling approach, and to incorporate the crime motivation, we need to narrow down the scale of our virtual world to study the interaction of agents among themselves and with other socio-economic, cultural, or environmental variables. According to RAT fundamentals, the model only considers rest, work, or leisure for agent's mandatory activities. The exclusion of either or the addition of other mandatory activities for particular demographics (e.g., senior citizens, parents of school-age children) can improve the realism of daily routines.

Finally, in our work, the model explainability was a priority to generate trust among LEAs, other institutions, specialists, and stakeholders. The lack of explainability in the models and techniques generally used in the predictive policing has raised ethical concerns in the US and calls into question their application in European contexts to fight crime [59]. For this reason, our decision model is inherently interpretable and encourages LEAs to trust it and use it correctly and effectively.

## IX. Conclusions and future work

This paper outlines the design and the theoretical foundations of a DS platform where human expertise and data-driven techniques have been combined into a DABMS approach to faithfully reproduce the dynamics of crime in an urban environment. The platform integrates the principles of the main criminal theories, such as RAT, RCT, CPT, Deterrence Theory, or the Repeat Victimization; it has been also instanced, calibrated, and evaluated in terms of performance with real data from the city of Málaga, a coastal city in southern Spain. The validation process has demonstrated its ability to provide accurate crime predictions, achieving a PAI and PEI at 3% of 12.41 and 0.93 respectively. The resulting DS can serve as a testing ground for LEAs to explore various crime prevention strategies, including police unit deployment tactics, and analyze their potential impact. This also allows authorities, policy makers, sociologists, or criminologists to assess complex urban scenarios that would be impractical to evaluate in real-world settings, often due to high costs and feasibility constraints. Our approach is characterized by the simplicity of the function used to estimate the number of crimes and the ease and explainability of the entire process, which enables, for instance, the swift allocation of police units across numerous deployment points, exceeding three thousand. This efficiency is achieved through a function relying on heuristics derived from historical data. Consequently, our DS could facilitate studies such as the analysis of the urban crime dynamics or even the exploration of urban police patrol allocation issues, allowing for empirical analysis of the effects of hotspot policing on crime reduction and prevention. Through our DS, various scenarios of police unit deployment can be explored to develop transparent strategies that have the potential to enhance the current distribution of law enforcement resources. The DS platform is designed to be continuously fed with historical crime data and therefore automatically calibrated from time to time to update its parameters.

Note also that our DS has the potential to become a DT if fed with real-time input data so that the physical world and the virtual space can interact each other in both directions. This would allow a closer analysis of the crime evolution and would permit LEAs to decide dynamically on the patrolling strategies of police units. We would also like to instantiate our platform in another city to confirm its suitability for modeling the urban crime. In the near future, we will use the platform to study strategies for partitioning the urban space in terms of crimes committed. The goal is to facilitate the LEAs' ability to optimize their staff. In this line, we will explore different techniques for deciding the best patrolling routes for certain urban sectors. Furthermore, we will analyze what factors could directly influence each type of crime, as well as what risks or benefits the potential aggressor may find upon executing a criminal act depending on the urban context in which he/she is located (period of time, characteristics of the place, meteorological factors, presence or absence of police officers, etc.). Finally, we want to enrich our platform with a model of human mobility based on real data. Several researchers [38] [65] have argued that street networks can have an impact on human activity and, as a consequence, on the pattern of crime events. Data sources that characterize mobility in an urban environment are not, in general, publicly available. Therefore, it is necessary to build models that, indirectly, provide evidence to predict, with relative accuracy, the parts of the city most frequented by citizens.

## ACKNOWLEDGMENT

This research is partially supported by the Spanish Ministry of Science and Innovation and by the European Regional *Development Fund (FEDER), Junta de Andalucía, and* Universidad de Málaga through the research projects with reference PID2021-122381OB-I00 and UMA20-FEDERJA-065.

The authors want to express special thanks to the Territorial Intelligence-Analysis Group of the National Police Force (*Unidad Territorial de Inteligencia del Cuerpo Nacional de Policía*, UTI-CNP) and to the main responsible for this LEA in the province of Málaga for their support and advice for this work.